%% file: main.tex
\documentclass[10pt,twocolumn,letterpaper]{article}

\usepackage{cvpr}              %

\input{preamble}

\definecolor{cvprblue}{rgb}{0.21,0.49,0.74}
\usepackage[pagebackref,breaklinks,colorlinks,allcolors=cvprblue]{hyperref}
\usepackage{float}
\def\paperID{4127} %
\def\confName{CVPR}
\def\confYear{2026}

\newcommand{\tipsvtwo}{TIPSv2\xspace}

\input{sec/00_title_author}

\begin{document}
\maketitle
\renewcommand{\thefootnote}{\fnsymbol{footnote}}
\footnotetext[1]{Authors contributed equally.}
\footnotetext[2]{Kaifeng Chen is now with xAI, Arjun Karpur with Epsilon Health, Bohyung Han with Seoul National University and Washington Ramos with Google. \\ Correspondence: \texttt{$\{$bingyi,koert,andrearaujo$\}$@google.com}}
\input{sec/0_abstract}    
\input{sec/1_intro}

\input{sec/2_rw}

\input{sec/3_panda}

\input{sec/5_results}

\input{sec/6_conclusion}
\input{sec/7_supp_mat_for_arxiv}
{
    \small
    \bibliographystyle{ieeenat_fullname}
    \bibliography{main}
}
\clearpage

\end{document}

%% file: preamble.tex
\usepackage{multirow}

\usepackage[table]{xcolor}
\usepackage{color}
\usepackage{xcolor}
\definecolor{citecolor}{rgb}{0.21,0.49,0.74}
\usepackage{colortbl}
\usepackage{array}
\definecolor{tblue}{RGB}{80,80,245}
\definecolor{tred}{RGB}{250,100,100}
\definecolor{tgreen}{RGB}{32,178,170}
\definecolor{rowred}{RGB}{240,230,230}
\usepackage{soul}
\sethlcolor{rowred}
\definecolor{ggplotgreen}{HTML}{00BFC4}
\definecolor{ggplotred}{HTML}{F8766D}
\usepackage{twemojis}

%% file: sec/00_title_author.tex
\title{{\textbf{\tipsvtwo}}:\\ \textit{Advancing Vision-Language Pretraining with Enhanced Patch-Text Alignment}}

\author{
\centerline{
Bingyi Cao\footnotemark[1]\hspace{8pt}  
Koert Chen\footnotemark[1]\hspace{8pt} 
Kevis-Kokitsi Maninis\hspace{8pt} 
Kaifeng Chen\footnotemark[2]\hspace{8pt} 
Arjun Karpur\footnotemark[2]}
\and\centerline{
Ye Xia\hspace{8pt}
Sahil Dua\hspace{8pt} 
Tanmaya Dabral\hspace{8pt}
Guangxing Han\hspace{8pt}
Bohyung Han\footnotemark[2]\hspace{8pt}
Joshua Ainslie
}
\and\centerline{
Alex Bewley\hspace{8pt}
Mithun Jacob\hspace{8pt} 
René Wagner\hspace{8pt} 
Washington Ramos\footnotemark[2]\hspace{8pt}
Krzysztof Choromanski
}
\and\centerline{
Mojtaba Seyedhosseini\hspace{8pt} 
Howard Zhou\hspace{8pt}
André Araujo}
\and\centerline{Google DeepMind} 
}

%% file: sec/0_abstract.tex
\begin{abstract}
Recent progress in vision-language pretraining has enabled significant improvements to many downstream computer vision applications, such as classification, retrieval, segmentation and depth prediction. However, a fundamental capability that these models still struggle with is aligning dense patch representations with text embeddings of corresponding concepts.
In this work, we investigate this critical issue and propose novel techniques to enhance this capability in foundational vision-language models.
First, we reveal that a patch-level distillation procedure significantly boosts dense patch-text alignment -- surprisingly, the patch-text alignment of the distilled student model strongly surpasses that of the teacher model.
This observation inspires us to consider modifications to pretraining recipes, leading us to propose iBOT++, an upgrade to the commonly-used iBOT masked image objective, where unmasked tokens also contribute directly to the loss.
This dramatically enhances patch-text alignment of pretrained models.
Additionally, to improve vision-language pretraining efficiency and effectiveness, we modify the exponential moving average setup in the learning recipe, and introduce a caption sampling strategy to benefit from synthetic captions at different granularities.
Combining these components, we develop \tipsvtwo, a new family of image-text encoder models suitable for a wide range of downstream applications. 
Through comprehensive experiments on 9 tasks and 20 datasets, we demonstrate strong performance, generally on par with or better than recent vision encoder models. 
Code and models are released via our project page at \url{https://gdm-tipsv2.github.io/}.

\end{abstract}
\vspace{-13pt}

%% file: sec/1_intro.tex
\section{Introduction}
\begin{figure}[t]
\begin{center}
   \includegraphics[width=1.0\linewidth]{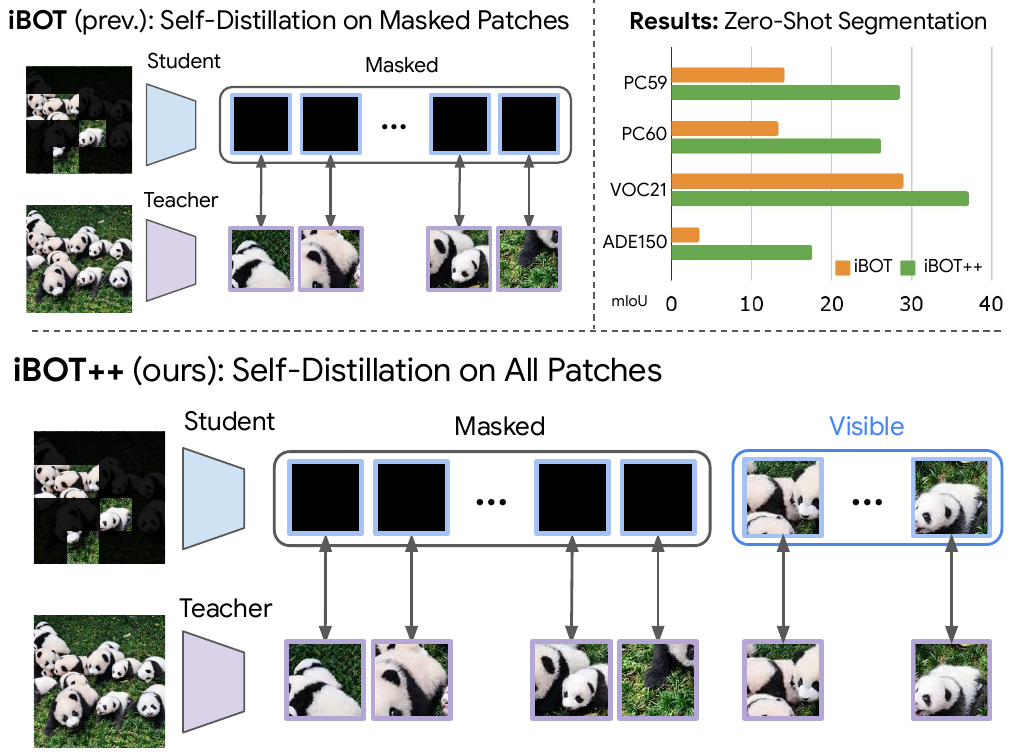}
\end{center}
\vspace{-0.1in}
  \caption{
  \textbf{\tipsvtwo's improvement to the masked image modeling pretraining strategy.}
  As part of our complete \tipsvtwo method, we introduce iBOT++ (bottom), a simple modification to the well-known iBOT~\cite{zhou2022ibot} self-supervised objective (top-left), where visible tokens also contribute directly to the loss.
  This enhancement dramatically improves patch-text alignment, as demonstrated by zero-shot image segmentation results (top-right).
 }
\label{fig:teaser_compact}
\end{figure}

Large-scale self-supervised and vision-language pretraining methods have fundamentally reshaped visual representation learning, establishing new performance standards across a broad spectrum of tasks. 
Leading approaches generally fall into two categories: weakly-supervised contrastive or sigmoid methods, exemplified by CLIP~\citep{radford2021clip}, SigLIP ~\citep{zhai2023siglip} and Perception Encoder (PE)~\citep{bolya2025perceptionencoderbestvisual}, which provide robust image-text alignment and zero-shot capabilities; and self-supervised learning (SSL) methods, such as DINO~\citep{siméoni2025dinov3,caron2021dino,oquab2024dinov2} and iBOT~\citep{zhou2022ibot}, which excel at spatial understanding for dense downstream tasks (e.g., depth, segmentation).

Achieving a unified representation that excels simultaneously at global (image-level) and dense (patch-level) understanding remains a significant challenge. State-of-the-art (SOTA) models often optimize one at the expense of the other: for example, DINOv2~\citep{oquab2024dinov2} excels at dense vision tasks but lacks inherent text alignment for vision-language capabilities, whereas PE-core~\citep{bolya2025perceptionencoderbestvisual} provides robust image-text alignment but underperforms on dense tasks. 
Recent unified approaches, such as TIPS~\citep{tips_paper} and SigLIP2~\citep{tschannen2025siglip2multilingualvisionlanguage}, have made significant strides in bridging this divide. 
However, we find that even these modern models struggle to maintain precise pixel- or patch-level alignment with text—a capability essential for advanced tasks like open-vocabulary segmentation. 
Self-supervised works such as DINO.txt~\citep{jose2024dinov2} and DINOv3~\citep{siméoni2025dinov3} attempt to address this by training supplementary text encoders, but still exhibit limited performance on multimodal tasks. 
Overall, grounded alignment with text remains a challenging task. 
This is corroborated by PE~\citep{bolya2025perceptionencoderbestvisual}, which finds that final transformer layers often function as global contrastive ``decoders'', rather than preserving the local semantics.

In this work, we first uncover a surprising trend in SOTA vision encoders ~\citep{tips_paper, tschannen2025siglip2multilingualvisionlanguage}: the largest flagship models often underperform their smaller counterparts in patch-text alignment. 
Investigating this issue with a recent spatially-aware multimodal encoder~\citep{tips_paper}, we find that distillation significantly improves grounding alignment by applying effective supervision across all patch tokens. 
This insight not only explains the performance gap, but also helps us unlock broader capabilities through a novel pretraining recipe. 

We introduce \textbf{\tipsvtwo}, the second version of the TIPS~\cite{tips_paper} model family (\textbf{T}ext-\textbf{I}mage \textbf{P}retraining with \textbf{S}patial awareness), a novel methodology designed to bridge this gap directly during pretraining.
\tipsvtwo incorporates a new objective, iBOT++, which translates our distillation findings into a self-supervised pretraining objective that enforces representation consistency on visible tokens, as per \cref{fig:teaser_compact}. 
We further enhance this with multi-granularity text augmentation (leveraging captions from PaliGemma ~\citep{beyer2024paligemma} and Gemini ~\citep{geminiteam2024gemini}) and a resource-efficient Exponential Moving Average (EMA) strategy. 
While standard SSL often requires full-model EMA, the presence of text supervision allows us to use a `head-only EMA'.
This updates only the projection layers, drastically reducing memory requirements at training time.
\tipsvtwo provides an efficient training framework yielding a highly effective image-text encoder across standard benchmarks. 
It sets new SOTA results in zero-shot semantic segmentation, while providing strong performance on a wide range of downstream multimodal tasks.
In summary, our primary contributions are:
\begin{itemize}
\item A novel finding that a distillation procedure tuned to strengthen spatial awareness unlocks strong gains in patch-text alignment.
Our detailed study pinpoints masking removal and initialization as the key factors driving this phenomenon.
\item iBOT++, a novel self-supervised masked objective that enhances pretraining to enforce representation consistency and significantly improves patch-text alignment -- see \cref{fig:teaser_compact}.
\item An efficient and effective training recipe leveraging multi-granularity text augmentation and a novel head-only EMA mechanism, overall achieving strong performance with significantly reduced training cost, validated comprehensively on 9 tasks spanning 20 datasets. 

\end{itemize}

%% file: sec/2_rw.tex
\section{Related Work}
\label{relatedwork}

There are currently three dominant paradigms for large-scale pretraining of versatile image representations: self-supervised visual learning~\cite{chen2020simclr,assran2023ijepa,he2022mae}, image-text contrastive learning~\cite{radford2021clip,jia2021align,cherti2023reproducible} or image captioning~\cite{tschannen2023cappa,tschannen2024locca,fini2025aimv2}.
In this work, we build on top of self-supervised and image-text contrastive techniques to construct visual representations that are natively text-aligned and spatially aware.
We review relevant work in these areas and contextualize our advances with respect to them.

\noindent\textbf{Self-supervised learning} methods enable visual pretraining without the use of any labels, by formulating pretext tasks that derive supervisory signals from the images themselves.
Landmark approaches in this area, such as DINO~\cite{caron2021dino} and iBOT~\cite{zhou2022ibot}, enforce representational invariances to image augmentations such as crops, flips, color distortions and masking.
These methods adopt a student-teacher distillation setup, where the teacher model is constructed as an exponential moving average (EMA) of the student~\cite{he2020moco}.
DINOv2~\cite{oquab2024dinov2} combines these strategies into a well-tuned recipe which was scaled on a curated training set.
More recently, DINOv3~\cite{siméoni2025dinov3} and WebSSL~\cite{fan2025scalinglanguagefreevisualrepresentation} scale this recipe even further to image encoders with $7$B parameters, on $2$B curated images.

Our proposed \tipsvtwo framework leverages self-supervised losses, making two contributions to enhance them.
First, we extend the iBOT~\cite{zhou2022ibot} masked image modeling (MIM) strategy to introduce our novel iBOT++ technique, which enables direct supervision on both visible and masked tokens at pretraining time -- this simple change dramatically boosts patch-text alignment and benefits several downstream tasks.
Examples in the literature have some similarities to the iBOT++ strategy.
DMAE~\cite{wu2023dmae} predicts visible and masked patches (not tokens), focusing on robust classifiers.
SimMIM~\cite{xie2022simmim} also attempted predicting visible patches, but found it ineffective.
MaskAlign~\cite{xue2023maskalign} and MR-MAE~\cite{gao2023mrmae} align student visible tokens with those of pretrained frozen models.
To the best of our knowledge, overall, supervising both visible and masked tokens as part of MIM pretraining was unexplored.
Second, we introduce a significant simplification to the widely-used EMA learning setup~\cite{he2020moco}, where only projection heads undergo EMA updates and a single vision encoder is used.
This is particularly effective in our setup because the additional image-text contrastive loss generally provides a stable signal preventing model collapse.
This reduces the number of trainable parameters by nearly half, enabling much more efficient scaling.
Other methods which combine SSL and contrastive image-text techniques are SILC~\cite{naeem2024silc}, TIPS~\cite{tips_paper} and SigLIP2~\cite{tschannen2025siglip2multilingualvisionlanguage}: all of them employ EMA teachers but do not consider simplifications in this component.

\noindent\textbf{Contrastive vision-language pretraining} was introduced by CLIP~\cite{radford2021clip} and ALIGN~\cite{jia2021align}, leveraging web-scale image datasets with weakly supervised text labels.
More recent instantiations in this area include OpenCLIP~\citep{cherti2023reproducible}, SigLIP~\citep{zhai2023siglip}, Perception Encoder~\cite{bolya2025perceptionencoderbestvisual}, EVA~\citep{fang2023eva,fang2024eva02,sun2023evaclip} and MetaCLIP~\cite{xu2024metaclip,chuang2025metaclip2}.
Some methods in this area~\cite{lai2024veclip,fan2023laclip,tips_paper,stone2025visualcomposition} explore the use of synthetic image captions for contrastive image-text training.
In our \tipsvtwo model, we go beyond these to combine noisy web captions with synthetic captions at different granularities, providing a range of possible textual descriptions for images, increasing the robustness of the model.

Knowledge distillation~\cite{hinton2015distilling} techniques have been frequently used to train smaller vision-language encoders~\cite{tips_paper,tschannen2025siglip2multilingualvisionlanguage,bolya2025perceptionencoderbestvisual}, enabling high performance at a much cheaper compute budget.
In this work, one of our contributions is to show for the first time that distillation procedures, tuned to strengthen spatial awareness, can be helpful to enhance patch-text alignment of these models.
Surprisingly, even large pretrained models with weak patch-text alignment can be distilled into smaller models to enhance this capability.

%% file: sec/3_panda.tex
\section{\textbf{\tipsvtwo}} \label{sec:method}

\subsection{Preliminaries} \label{sec:preliminaries}

Our method builds on top of the previous TIPS~\cite{tips_paper} version, a recent vision encoder pretraining method that integrates contrastive image-text learning with self-supervised learning.
In this section, we review the necessary background and introduce the notation used throughout the paper.
Given a collection of image-text pairs $\{(I_k,T_k)\}$, we learn an image encoder $f$ mapping an image $I$ to image embeddings $\{\mathbf e^g,\mathbf e_1,\mathbf e_2,\ldots,\mathbf e_N\}$
and a text encoder $g$ mapping texts $T$ to a text embedding $\mathbf e^t$. Let $f^g(I)=\mathbf e^g$ be the global embedding representation of the entire image and $f(I)=\{\mathbf e_1,\mathbf e_2,\ldots,\mathbf e_N\}$ the patch embeddings corresponding to different image regions. 
$f$ is modeled as a Vision Transformer (ViT)~\citep{dosovitskiy2020vit} and $g$ is modeled as a standard transformer \citep{vaswani2017attention}. The loss is given by $\mathcal{L} = \mathcal{L}_\text{CLIP} + \mathcal{L}_\text{DINO} + \mathcal{L}_\text{iBOT}$, and each component is detailed below.

\noindent\textbf{Image-text contrastive loss.} The first component, $\mathcal{L}_\text{CLIP}$, aligns image and text modalities. Following the approach of CLIP~\citep{radford2021clip}, $\mathcal{L}_\text{CLIP}$ is an InfoNCE~\citep{oord2018infonce} loss that pushes $\mathbf e^g$ and $\mathbf e^t$ close for corresponding images and captions, and far otherwise. TIPS adapts this by introducing two separate CLS global embeddings: one supervised by web alt-text captions for object-centric details, and another supervised by PaliGemma~\citep{beyer2024paligemma} synthetic captions designed to better capture dense spatial relationships. 

\noindent\textbf{Global- and patch-level SSL.}
In addition, TIPS applies two forms of self-supervised losses to boost spatial awareness, using a teacher network $f_t$ that is the exponential moving average (EMA) of the student network $f_s$. In these losses, the teacher network $f_t$ processes the original image $I$, and the student network $f_s$ processes a crop or masked view. Then, these embeddings are projected into a higher ``prototype" dimension by head networks $h_s$ and $h_t$, and the resulting prototypes are supervised to be invariant to these transformations by a cross-entropy loss.
In the global-level self-distillation objective (DINO~\citep{caron2021dino}), the student representations of $M$ local crops $\{I_i\}_{i=1}^M$ are compared to the teacher representation of $I$ on global embeddings:
\begin{equation}
    \label{eq:dino_loss}
        \mathcal{L}_\text{DINO} = - \sum_{i=1}^M  h_t(f_t^g(I))^T \log h_s(f_s^g(I_i))
\end{equation}

\input{tables/tab_zeroseg_scale}

\noindent In the patch-level MIM objective (iBOT~\citep{zhou2022ibot}), a random binary mask $m$ is applied to image patches to obtain masked view $I_\text{mask}$ ($m_i=1$ denotes that the $i$-th patch is masked). Then, the student representation of $I_\text{mask}$ is compared to the teacher representation of $I$ on patch embeddings: 
\begin{equation}
    \label{eq:ibot_loss}
        \mathcal{L}_\text{iBOT} = - \sum_{i=1}^N  m_i  h_t(f_t(I)_i)^T \log h_s(f_s(I_\text{mask})_i)
\end{equation}

\begin{figure*}[ht]
    \centering
    \begin{minipage}[c]{0.70\linewidth}
        \vspace{-1mm}
        \scriptsize
        \centering
        {
            \resizebox{1.05\linewidth}{!}{
                \input{tables/tab_init}
            }
        }
        \captionof{table}{\textbf{Initialization and masking ablations for distillation.} Comparing TIPS ViT-L models re-trained with different strategies. We highlight \textbf{best} and \underline{second-best} overall. Initialization is either from scratch (``Random") or initialized from the teacher model (``Pretrained"). Update is either training (\twemoji{fire}) or frozen (\twemoji{snowflake}). The results show that removing masking and randomly initializing the student image encoder are critical for achieving strong patch-text alignment.
        }
        \label{tab:init}
    \end{minipage}
    \hfill %
    \begin{minipage}[c]{0.25\linewidth}
        \centering
        \includegraphics[width=\linewidth]{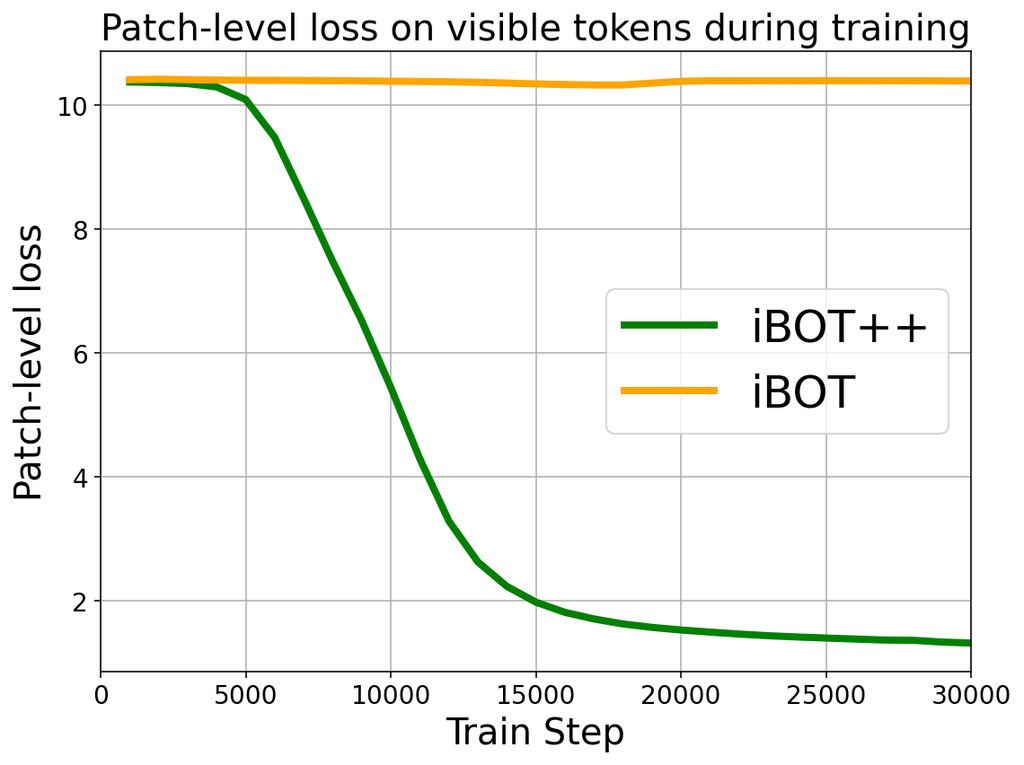}
        \vspace{-6.5mm}
        \caption{
        The decrease in patch-level loss for visible tokens using iBOT++ indicates successful anchoring to the teacher's tokens, which does not happen for iBOT.
        }
        \label{fig:ibot++}
    \end{minipage}
\end{figure*}

\noindent\textbf{Distillation.}
TIPS also distills a family of smaller models with a simple strategy that mirrors the pretraining setup. Distillation only requires two modifications: the teacher network $f_t$ is replaced with a frozen larger teacher instead of an EMA copy of $f_s$, and no mask is applied at the patch-level loss. 

\subsection{Bridging Pretraining and Distillation}
\label{bridge}

In this section, we study the effect of distillation techniques on patch-text alignment. Our main finding is that the patch-level distillation procedure can significantly enhance patch-text alignment, even when utilizing a teacher with poor alignment. To demonstrate this, we begin by evaluating patch-text alignment using standard zero-shot semantic segmentation benchmarks: ADE150~\citep{zhou2017ade20k, zhou2018semanticunderstandingscenesade20k}, Pascal Context (PC59/PC60)~\citep{Mottaghi_2014_CVPR}, and Pascal VOC (VOC21)~\citep{everingham2010pascal}.

As shown in \cref{tab:zeroseg_scale}, the largest pretrained ViT-g TIPS model significantly underperforms the smaller ViT-L student model (which is distilled from the ViT-g) in the task of zero-shot segmentation, reversing the trend of all other tasks reported in \citep{tips_paper}. 
This discrepancy suggests that while the pre-training recipe at the ViT-g scale is highly effective for image-only or global tasks, it fails to induce strong patch-text alignment, a capability apparently gained during distillation. 
To disentangle the factors distinguishing pretraining from distillation, \cref{tab:init} presents ablations using a ViT-L self-distillation baseline. We systematically vary the masking ratio and model initialization/freezing parameters to isolate their individual contributions:
\begin{itemize} 
\item  \textbf{Distillation method:} The models in rows (2-7) are distilled from the frozen ViT-L teacher trained in row (1). These follow the distillation procedure of \citep{tips_paper} described in \cref{sec:preliminaries}, except that the teacher is the same size as the student, to study the setting of distillation in isolation from the effect of a larger teacher. The other differences from vanilla distillation are described row-by-row.

\item \textbf{Masking ratio:} We ablate the effect of masking ratio by lowering it through rows (2-4). In (2), the ratio is the same as vanilla pre-training, and in (4) it is lowered to no masking, which is the ratio in vanilla distillation. Observe that row (2) is the pre-training setting with two key changes: the teacher is frozen and, more importantly, the patch-level loss is also applied to the $25\%$ of unmasked tokens instead of unsupervised as in pre-training. This configuration already yields a significant enhancement in zero-shot segmentation, demonstrating the potential of patch-level distillation for improving patch-text alignment. In contrast, observe that (4) exactly matches the vanilla distillation of \citep{tips_paper}. The increase in patch-text alignment from rows (2-4) as we lower the masking ratio indicates that the loss on visible tokens is critical for achieving improved patch-text alignment during distillation.

\item \textbf{Initialization and freezing:} We ablate the effect of random vs. pre-trained initialization and freezing the text encoder in the remaining rows. We find that the text encoder is less sensitive to initialization in rows (4–6). However, row (7) demonstrates that random initialization is essential for the vision encoder. Initializing distillation with the pre-trained visual encoder as in row (7) completely eliminates the advantage from distillation, reducing it to the performance of the original pre-trained model. This suggests that the model must break away from the pretrained convergence region to learn effectively during distillation.

\end{itemize}

\begin{figure}[t]
    \centering
    \includegraphics[width=1.0\linewidth]{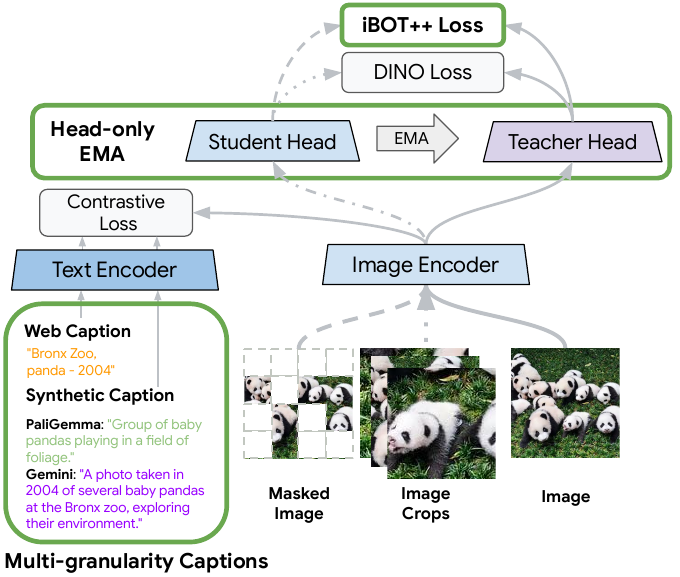}
    \caption{
    \textbf{\tipsvtwo pretraining overview}. \tipsvtwo introduces $3$ improvements (highlighted in green borders) to the combined contrastive and self-supervised approach to pretrain vision encoders.
    iBOT++ is an enhanced masked image modeling loss.
    Head-only EMA enables memory-efficient self-supervised losses.
    Multi-granularity captions provide a range of possible textual descriptions for images, increasing the robustness of the model.    
    }
    \label{fig:block}
\end{figure}

These findings suggest that supervising all patch tokens, rather than just masked ones, is crucial for alignment during distillation. We leverage these insights to formulate a novel pretraining strategy, to be discussed next.
An overview of the final \tipsvtwo pretraining recipe is presented in \cref{fig:block}.

\subsection{iBOT++}
\label{loss_diff}

As discussed in \cref{sec:preliminaries}, both the pretraining and distillation stages of the vision encoder reference method incorporate a patch-level loss.
The pretraining phase employs the iBOT \citep{zhou2022ibot} MIM objective, where the student reconstructs masked tokens by matching the features of a teacher encoder that processes the unmasked image.
For distillation, masking is removed entirely, so both the student and teacher see the unmasked image.

Based on \cref{bridge}, we hypothesize that limiting loss to masked regions discourages preserving local semantics. To resolve this, we propose \textbf{iBOT++}, a modification to iBOT which applies the patch-level loss to all patches, both masked and unmasked.
This simple change represents a significant improvement to the pretraining stage of TIPS.
The improved loss $\mathcal{L}_\text{iBOT++}$  differs from the original $\mathcal{L}_\text{iBOT}$ in \cref{eq:ibot_loss} only in that the loss also applies on visible tokens: 
\begin{equation}
    \label{eq:ibot_plus_loss}
        \mathcal{L}_\text{iBOT++} = - \sum_{i=1}^N   h_t(f_t(I)_i)^T \log h_s(f_s(I_\text{mask})_i)
\end{equation}

While foundational vision encoders~\citep{tips_paper, oquab2024dinov2, siméoni2025dinov3} require high masking ratios (e.g., $75\%$) for successful MIM~\citep{he2022mae} in pre-training, we found in \cref{bridge} that distillation instead benefits substantially from matching student to teacher representations across all tokens without masking. 
iBOT++ operates as an intermediate between these two objectives: we retain the masking mechanism but extend the supervision to visible tokens, requiring the student to align with teacher representations for both masked and unmasked patches. 
Replacing iBOT with iBOT++ in TIPS pretraining yields significant gains in zero-shot segmentation (\cref{tab:fromscratch_distill}), indicating superior patch-text alignment. 
\cref{results} demonstrates that iBOT++ also generally benefits other tasks.

Note that standard iBOT lacks direct supervision on visible tokens, which allows their representations to behave arbitrarily provided they suffice for reconstructing masked tokens. 
Enforcing alignment on unmasked tokens enables iBOT++ to anchor student patch representations to the teacher's.
This is demonstrated in \cref{fig:ibot++}, which indicates through the proxy of patch-level loss that the representations of visible patches remain poor in iBOT but improve in iBOT++.
Overall, iBOT++ delivers superior performance by simultaneously capturing global context for reconstruction and preserving local semantics.

\subsection{Head-only EMA}
\label{init_diff}

Standard self-distillation ~\citep{caron2021dino, zhou2022ibot, oquab2024dinov2, tips_paper} relies on an EMA teacher network to provide stable targets via temporal ensembling.
Crucially, in the context of SSL-only methods, an EMA teacher with a stop gradient avoids~\citep{grill2020bootstrap} the trivial solution where the prototype representations of student $h_s(f_s(I))$ and teacher $h_t(f_t(I))$ collapse to a constant representation regardless of input $I$. %
This collapse can happen at either the encoder $f_s, f_t$ or at the head $h_s, h_t$.
However, in the context of combined SSL and image-text contrastive learning, the encoders $f_s$ and $f_t$ are already constrained to prevent collapse by the contrastive loss $\mathcal{L}_\text{CLIP}$.
As a result, we propose that it is sufficient to use the EMA teacher only for the head network, and entirely eschew the EMA teacher for the main vision encoder.
That is, we set $f_t = f_s$ and continue only to update $h_t$ by EMA updates from $h_s$. 

Ablations in \cref{results} demonstrate that applying EMA exclusively to the projector head retains the majority of the performance of applying EMA to both head and encoder while significantly reducing resource requirements.
As EMA methods require a second copy of all parameters being averaged during training, this resource-efficient adaptation represents significant savings in memory overhead and training throughput.
For example, on ViT-B, this method reduces training parameters by $42\%$.
Also, note that this design represents a necessary intermediate, as our preliminary experiments find that entirely removing EMA by sharing both the encoder and projector causes severe training instability and performance degradation.

\input{tables/tab_fromscratch_ablation}

\begin{figure}[t]
\begin{center}
   \includegraphics[width=1.0\linewidth]{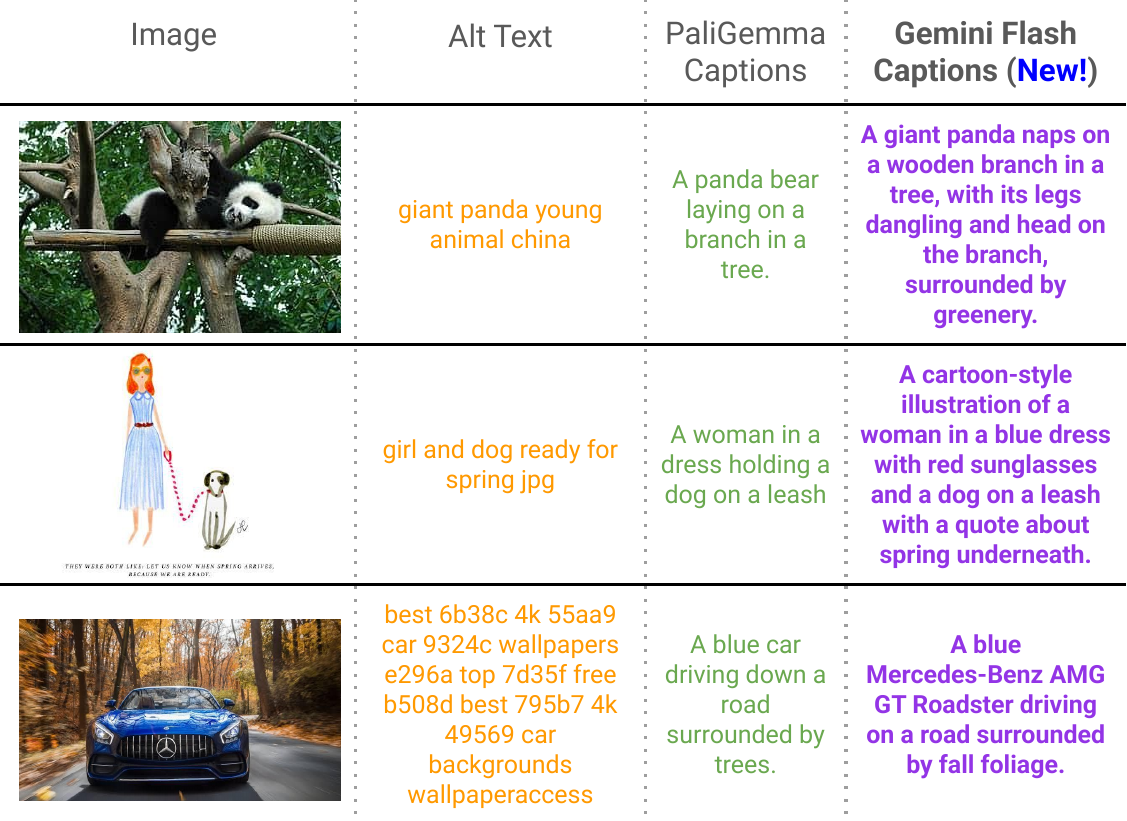}
\end{center}
\vspace{-2mm}
  \caption{
  \textbf{Image captions at different granularities.}
  Besides the alt-text and PaliGemma captions used in previous work, we propose additionally leveraging the more comprehensive Gemini captions, to make our vision encoder more robust.
 }
\label{fig:captions}
\end{figure}

\subsection{Multi-Granularity Text Captions}
\label{multi_caption}

 \cref{fig:captions} illustrates limitations in the alt-text and PaliGemma captions used in~\citep{tips_paper}. 
 For instance, in the panda image, both captions fail to describe the animal's pose (dangling legs and head resting on the branch).
 Similarly, in the illustration example, neither caption identifies the image as a cartoon, a key piece of semantic information. 
 Finally, in the last example, both captions overlook the seasonal context, where the yellow leaves indicate autumn.
 As the vision encoder's learning is directly tied to caption quality, these failures motivate the need for richer, more descriptive captions.

\input{tables/tab_ablations}

We extend this by generating richer synthetic captions using Gemini Flash \citep{geminiteam2024gemini}, conditioned on the image, alt-text, and PaliGemma captions, as \cref{fig:captions}.
Despite their higher quality, preliminary experiments showed that longer captions underperformed in isolation.
This occurs because overly comprehensive captions trivialize the contrastive loss ($\mathcal{L}_\text{CLIP}$).
The abundance of detail facilitates discrimination within the batch, ultimately hindering the learning of robust representations.
To address this, we randomly alternate between detailed Gemini captions and simpler PaliGemma captions during training to supervise the second CLS token. 
This balances learning difficulty and detail extraction, empirically boosting both dense and global performance.

%% file: tables/tab_zeroseg_scale.tex
\begin{table}[t]
\caption{\small\textbf{Zero-shot segmentation with a large teacher model and its distilled student.} Surprisingly, the TIPS ViT-L model, which is a student distilled from the TIPS ViT-g teacher, significantly surpasses it.}
\vspace{-2mm}
\scriptsize
\centering
{
\begin{tabular}{l c c c c}
\small \multirow{2}{*}{Model} & \multicolumn{4}{c}{Zero-shot Segmentation (mIoU) $\uparrow$} \\
 & PC59 & PC60 & VOC21 & ADE150 \\
\midrule
TIPS ViT-L & \bf{33.5} & \bf{30.4} & \bf{30.5} & \bf{20.8} \\
TIPS ViT-g & 11.4 & 10.8 & 19.7 & 2.6 \\
\end{tabular}
}
\vspace{-0mm}
\label{tab:zeroseg_scale}
\end{table}

%% file: tables/tab_init.tex
\begin{tabular}{l r r c c c c c c}
\small \multirow{2}{*}{Setting} & \multicolumn{2}{c}{Initialization / Update} & Masking Ratio & \multicolumn{4}{c}{Zero-shot Segmentation (mIoU) $\uparrow$} \\
 & Text Encoder & Student Encoder & & PC59 & PC60 & VOC21 & ADE150  \\
\midrule
(1) pretraining & Random / \twemoji{fire}  & Random / \twemoji{fire} & 0.75 & 6.9 & 6.6 & 6.7 & 0.3 \\ 
(2) distillation & Random / \twemoji{fire}  & Random / \twemoji{fire} & 0.75 & 16.0 & 15.4 & 22.5 & 5.9 \\
(3) distillation & Random / \twemoji{fire}  & Random / \twemoji{fire} & 0.5 & 15.5 & 14.5 & 24.0 & 7.0 \\
(4) distillation & Random / \twemoji{fire}  & Random / \twemoji{fire} & 0.0 & \bf{31.4} & \bf{28.6} & \underline{30.8} & \underline{20.0} \\
(5) distillation & Pretrained / \twemoji{fire}  & Random / \twemoji{fire} & 0.0 & \underline{31.1} & \underline{28.5} & \bf{31.9} & \bf{20.5} \\
(6) distillation & Pretrained / \twemoji{snowflake} & Random / \twemoji{fire} & 0.0 & 30.1 & 27.7 & 28.4 & \underline{20.0} \\
(7) distillation & Pretrained / \twemoji{fire}  & Pretrained / \twemoji{fire} & 0.0 & 6.4 & 6.0 & 6.7 & 2.4 \\
\end{tabular}

%% file: tables/tab_fromscratch_ablation.tex
\begin{table}[t]
\caption{\small\textbf{Zero-shot segmentation in pretraining.} Comparing TIPS ViT-g with iBOT or iBOT++ methods, showing significant improvements with our novel iBOT++.}
\vspace{-1mm}
\scriptsize
\centering
{
\begin{tabular}{l c c c c}
\small \multirow{2}{*}{Model} & \multicolumn{4}{c}{Zero-shot Segmentation (mIoU) $\uparrow$} \\
& PC59 & PC60 & VOC21 & ADE150  \\
\midrule
TIPS~\cite{tips_paper} (with iBOT) & 14.2 & 13.4 & 29.1 & 3.5 \\ 
TIPS with iBOT++ & \bf{28.6} & \bf{26.2} & \bf{37.2} & \bf{17.6} \\
\end{tabular}
}
\vspace{-0mm}
\label{tab:fromscratch_distill}
\end{table}

%% file: tables/tab_ablations.tex
\begin{table*}[t]
\caption{\small\textbf{Ablation studies for \tipsvtwo's pretraining technique}. Ablations are cumulative, running on a fixed schedule of $100$k steps at resolution $224$, for fair comparisons. We highlight the \textbf{best} and \underline{second-best} number of each column.}
\vspace{-2mm}
\scriptsize
\addtolength{\tabcolsep}{0.2em}
\centering
{
\begin{tabular}{l c c c c c c c}
\small \multirow{2}{*}{Model} & Seg. $\uparrow$ & Depth $\downarrow$ & Normals $\downarrow$ & ImageNet $\uparrow$ & I$\rightarrow$T Ret. $\uparrow$ & T$\rightarrow$I Ret. $\uparrow$ & Zero-shot Seg. $\uparrow$ \\
& PASCAL & NYUv2 & NYUv2 & KNN & Flickr & Flickr & ADE150 \\
\midrule
TIPS ViT-g (reproduced)& 82.8 & 0.375 & 23.1 & 83.2 & 92.0 & 81.0 & 3.5 \\
+ iBOT++ & 82.5 & 0.369 & \textbf{22.7} & \textbf{84.4} & 93.9 & \underline{81.7} & 17.6 \\
+ Multi-granularity Captions & \underline{83.7} & \underline{0.354} & \textbf{22.7} & \underline{84.3} & \underline{95.0} & \textbf{85.4} & \underline{18.1} \\
+ Head-only EMA & \textbf{83.8} & \textbf{0.353} & \underline{22.8} & 84.1 & \textbf{95.4} & \textbf{85.4} & \textbf{19.1} \\
\end{tabular}
}
\label{tab:ablation_studies}
\end{table*}

%% file: sec/5_results.tex
\section{Results}
\label{results}
\subsection{Experimental Setup}

To demonstrate general-purpose capability as a vision encoder, we evaluate our models on a suite of 9 standard tasks, covering two broad types: image-text and image-only.
Datasets and metrics used in this work are introduced in the following paragraphs.
Evaluations follow the detailed protocol of \citep{tips_paper}, to ensure fair comparisons.
As our objective is to assess the applicability of our model off-the-shelf for downstream applications, the pretrained image-text representations are kept frozen for all evaluations.

\noindent\textbf{Image-text tasks} include 2 types, focused on dense and global image-text alignment.
We consider dense image-text alignment by evaluating on zero-shot segmentation, reporting mean Intersection over Union (mIoU).
Unless otherwise mentioned, we report \tipsvtwo results on zero-shot segmentation with the simple protocol that maximizes cosine similarities between patch features, up-scaled to image resolution, and the text features of each class name.
Note that DINOv2 (with text~\cite{jose2024dinov2}) and SILC~\cite{naeem2024silc} instead report results using the expensive sliding window protocol from TCL~\citep{cha2023learning}, which generally boosts results.
Both protocols use the value embeddings of the final transformer layer as patch embeddings, following ~\citep{zhou2022extract}.
We report zero-shot segmentation on ADE20K~\citep{zhou2017ade20k} with 150 classes (ADE150), PASCAL context \citep{everingham2011pascal}, and PASCAL VOC~\citep{everingham2010pascal}.
For PASCAL context, we report both with (PC60) and without (PC59) the background label.
The global-focused tasks include image-to-text and text-to-image retrieval on Flickr30K~\citep{young2014from}, DOCCI~\citep{onoe2024docci}, and COCO~\citep{chen2015cococaptions} reporting recall@1; and zero-shot classification on ImageNet-1K reporting top-1 accuracy by retrieving nearest class text embeddings~\citep{radford2021clip}.

\noindent\textbf{Image-only tasks} include semantic segmentation with linear probe on PASCAL VOC~\citep{everingham2010pascal} and ADE20k~\citep{zhou2017ade20k} reporting mIOU; monocular depth estimation on the scene-centric NYUv2~\citep{silberman2012indoor} with linear probe and the object-centric NAVI~\citep{jampani2023navi} with DPT~\cite{ranftl2021vision}, both reporting RMSE; surface normal estimation on NYUv2 and NAVI, using DPT, reporting angular RMSE; image classification on ImageNet-1K~\citep{russakovsky2015imagenet} reporting top-1 accuracy (KNN and linear probe); and fine-grained/instance-level retrieval on the Universal Embeddings Dataset (UnED)~\citep{ypsilantis2023uned} (comprising data from $8$ different domains: Food2k \citep{min2023large}, CARS196 \citep{krause20133d}, SOP \citep{song2016deep}, InShop \citep{liu2016deepfashion}, iNat \citep{van2018inaturalist}, Met\citep{ypsilantis2021met}, GLDv2 \cite{weyand2020google}, Rp2k\cite{peng2020rp2k}), reporting average recall@1.

\noindent\textbf{Training data.}
We train our models using the web-crawled WebLI dataset~\citep{chen23pali}, reusing the specific filtered subset created by~\cite{tips_paper}, containing $116$M images.
We also reuse synthetic captions from PaliGemma~\citep{beyer2024paligemma}, again following~\cite{tips_paper}.
Additionally, as described in \cref{multi_caption}, we leverage synthetic captions from Gemini 1.5 Flash \cite{geminiteam2024gemini}.

\noindent\textbf{Implementation details.}
Following \citep{tips_paper}, pretraining combines contrastive image-text and self-supervised losses, integrating the key upgrades described in \cref{sec:method}: iBOT++, head-only EMA and multi-granularity text captions.
Pretraining consists of a low-resolution stage for $90k$ steps at batch size $8192$ and then a high-resolution adaptation stage for $9k$ steps at batch size $4096$.
At low resolution, we train with $1$ global crop at resolution $224$ and $6$ local crops at resolution $98$.
At high resolution, the global crop has resolution $448$ and the local crops have resolution $140$.
We use only random resizes, crops and flips as augmentations.
We pre-train at ViT-g scale with 512 TPUv5 chips for 2 days.
We also distill to a family of -SO, -L, -B models, with all results presented in the appendix.
Distillation is as described in \cref{sec:preliminaries}, and then student models also undergo a high-resolution adaptation stage.

\subsection{Ablations}
We present ablations in \cref{tab:ablation_studies} where we add each of the primary contributions successively, evaluated on a representative subset of 7 of our evaluation tasks.
We begin with TIPS~\citep{tips_paper} as a baseline, reproduced with two minor modifications: we halve the batch size and increase position embeddings to $32\times32$ from $16\times16$.
Our preliminary experiments showed halving the batch size saved training resources without affecting results significantly, and higher-resolution position embeddings improved all results without incurring much cost. 
Ablations are carried out for 100k steps at original resolution, without the high-res adaptation stage of the full pretraining process.
In the second row, iBOT is replaced by iBOT++, which results in across-the-board improvements for most evals, including a dramatic $+14.1$ mIoU improvement to zero-shot segmentation.
After that, we replace PaliGemma synthetic captions with multi-granularity captions sampled from PaliGemma and Gemini captions, which improves both global and dense types of image-text tasks: I$\rightarrow$T, T$\rightarrow$I, and zero-shot segmentation.
Finally, we add head-only EMA, a change that is primarily targeted at enabling significant training resource savings, but still results in comparable performance, even further improving some tasks such as  zero-shot segmentation. 
The combination of these changes is \tipsvtwo, with all primary components resulting in increased or comparable performance across our evaluations, with particular significant improvement for zero-shot segmentation from iBOT++.

\input{tables/tab_sota_dense_image_text}
\input{tables/tab_sota_image_text} 
\input{tables/tab_sota_image_only}
\input{tables/tab_dinov3}

\subsection{Comparisons with Other Vision Encoders}

We provide extensive benchmarking against state-of-the-art methods.
For fair comparisons, we strictly follow the protocol of \citep{tips_paper}, considering the largest instantiations in each model family, up to ViT size ``G", with $1.8$B parameters in the vision encoder.
Note that the largest model in the \tipsvtwo family is of size ``g", with $1.1$B parameters in the vision encoder.
The compared methods are the following.
Weakly-supervised methods: CLIP~\citep{radford2021clip}, PE~\citep{bolya2025perceptionencoderbestvisual}, and OpenCLIP~\citep{cherti2023reproducible}; self-supervised methods: DINOv2~\citep{oquab2024dinov2} (with registers~\cite{vitsneedregisters}), DINOv3~\citep{siméoni2025dinov3}, and FRANCA~\citep{venkataramanan2025franca}; combined approaches: TIPS~\citep{tips_paper}, SILC~\citep{naeem2024silc}, dino.txt~\citep{jose2024dinov2} (as the image-text capability of DINOv2), and SigLIP2~\citep{tschannen2025siglip2multilingualvisionlanguage}.
Qualitatively, \cref{fig:qualitative_dense} shows that \tipsvtwo patch embeddings are more spatially coherent than other vision-language pretraining methods.

\noindent\textbf{Dense image-text tasks.} In \cref{tab:sota_dense_image_text}, we report zero-shot segmentation results on ViT-L and the closest comparable sizes from competitor models.
\tipsvtwo achieves state-of-the-art results, indicating a significant improvement in dense patch-text alignment. 
Qualitative visualizations of the zero-shot segmentation results are presented in \cref{fig:zss_viz}.

\noindent\textbf{Global image-text tasks.} 
In \cref{tab:sota_image_text}, \tipsvtwo achieves the best or second-best results in 5 out of 7 evaluations. Our models are best in all COCO and long-text DOCCI tasks, outperforming even larger ViT-G models.
Note that our ViT-g models even outperform the recent PE~\citep{bolya2025perceptionencoderbestvisual} ViT-G models in 3 out of 5 evals, although the PE model has 56\% more parameters and processes 47$\times$ more image-text pairs.

\newcommand{\showpca}[2]{%
  \resizebox{#1\linewidth}{!}{%
	\setlength{\fboxsep}{0pt}
		\hfill
		\includegraphics[height=#1\linewidth]{images/#2.jpeg}
		\hfill
		\includegraphics[height=#1\linewidth]{images/tips/#2.png}
		\hfill
		\includegraphics[height=#1\linewidth]{images/siglip2/#2.png}
		\hfill
		\includegraphics[height=#1\linewidth]{images/supp/tipsv2/vit_g/#2.png}
}}

\newcommand{\pcawidth}{1.0}
\begin{figure}[tb]
	\centering
	\showpca{\pcawidth}{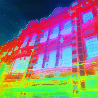} \\[0.2mm]
    \showpca{\pcawidth}{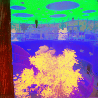} \\[0.2mm]
    \showpca{\pcawidth}{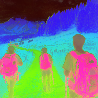} \\[0.2mm]
    \showpca{\pcawidth}{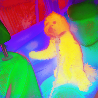} \\[0.1mm]

    \resizebox{\pcawidth\linewidth}{!}{%
    \scriptsize
        \begin{minipage}[t]{0.24\linewidth}
          \centering
          Image
        \end{minipage}%
        \hfill
        \begin{minipage}[t]{0.24\linewidth}
          \centering
          TIPS
        \end{minipage}%
        \hfill
        \begin{minipage}[t]{0.24\linewidth}
          \centering
          SigLIP2
        \end{minipage}%
        \hfill
        \begin{minipage}[t]{0.24\linewidth}
          \centering
          \tipsvtwo (ours)
        \end{minipage}%
    }%
    \vspace{-1mm}
	\caption{\small\textbf{PCA maps.} Comparing the first $3$ PCA components, using ViT-g models. 
    Images are forwarded at 1372 resolution for patch size 14 models (TIPS, TIPSv2) and at 1568 resolution for patch size 16 models (SigLIP2).
    \tipsvtwo produces smoother feature maps compared to other vision-language pretraining methods, with well-delineated objects.}
	\label{fig:qualitative_dense}
	\vspace{-3mm}
\end{figure}

\noindent\textbf{Image-only tasks.} In \cref{tab:sota_image_only}, \tipsvtwo achieves the best or second-best results in 7 out of 9 evaluations, and also achieves strong performance on the remaining 2 cases.
\tipsvtwo stands out as particularly strong in dense understanding tasks, improving by $+1.5$ in segmentation on PASCAL and by $-0.019$ on depth in NYUv2 over prior best.
However, our models do not perform as well in ImageNet classification, a result of our focus on optimizing for general purpose capability across a wide range of tasks.

\noindent\textbf{DINOv3 comparison.}
Additionally, we compare \tipsvtwo with the very recent DINOv3~\citep{siméoni2025dinov3}, presenting results in \cref{tab:dinov3}.
In order to make the zero-shot segmentation evals comparable, for this table we reproduce their setup, with the more expensive sliding window protocol from TCL~\citep{cha2023learning}.
Note that the DINOv3 teacher model is trained with 6$\times$ more parameters and 15$\times$ more images than the \tipsvtwo teacher, so we believe the fairest possible comparison is to use the largest model size in common between the two model families, which is ViT-L.
We report numbers on the subset of all released evals with matching protocols.
Notably, \tipsvtwo outperforms DINOv3 in 4 out of 6 evals, despite the data and model size disadvantage.

%% file: tables/tab_sota_dense_image_text.tex
\begin{table}
\caption{\small\textbf{Dense image-text evaluations}, where \tipsvtwo outperforms others in all cases, even though SILC and DINOv2 use the more expensive TCL protocol~\cite{cha2023learning}. We highlight the \textbf{best} and \underline{second-best} number of each column.
}
\vspace{-2mm}
\scriptsize
\centering
{
\begin{tabular}{l c c c c}
\small \multirow{2}{*}{Method} & \multicolumn{4}{c}{Zero-shot Segmentation $\uparrow$}\\
& PC59 & PC60 & VOC21 & ADE150  \\
\midrule
SigLIP2 SO/14 & - & 19.6 & 26.8 & 15.6 \\
SILC B/16 & 31.6 & - & - & 19.3 \\
DINOv2 (dino.txt) L/14 & 30.9 & - & - & 20.6 \\
TIPS L/14 & \underline{33.5} & \underline{30.4} & \underline{30.5} & \underline{20.8} \\
\tipsvtwo L/14 (ours) & \textbf{37.1} & \textbf{33.9} & \textbf{44.4} & \textbf{24.7} \\

\end{tabular}
}

\label{tab:sota_dense_image_text}
\end{table}

%% file: tables/tab_sota_image_text.tex
\begin{table*}[ht]
\caption{\small\textbf{Global image-text evaluations}, where \tipsvtwo achieves best or second-best in $5$ out of $7$ cases. We highlight the \textbf{best} and \underline{second-best} number of each column.}
\vspace{-2mm}
\scriptsize
\addtolength{\tabcolsep}{0.5em}
\centering
{
\begin{tabular}{l c c c c c c c c}
\small \multirow{2}{*}{Model} & \multicolumn{3}{c}{I$\rightarrow$T Retrieval $\uparrow$} & \multicolumn{3}{c}{T$\rightarrow$I Retrieval $\uparrow$} & \multicolumn{1}{c}{0-Shot $\uparrow$} \\
& COCO & Flickr & DOCCI & COCO & Flickr & DOCCI & ImageNet \\
\midrule
CLIP L/14 & 56.3 & 85.2 & 44.4 & 36.5 & 65.2 & 40.4 & 75.5 \\
OpenCLIP G/14 & 67.3 & 92.9 & - & 51.4 & 79.5 & - & 80.1 \\
SigLIP2 g/16 & 72.8 & \underline{95.4} & - & 56.1 & \textbf{86.0} & - & \underline{85.0} \\
SILC G/16 & 73.2 & - & - & 54.7 & - & - & 83.7 \\
DINOv2 (dino.txt) L/14 & - & - & - & 45.4 & 77.1 & - & 81.4 \\
PE-core G/14 & \underline{75.4} & \textbf{96.2} & - & 58.1 & 85.7 & - & \textbf{85.4} \\
TIPS g/14 & 74.0 & 93.0 & \underline{57.2} & \underline{59.4} & 84.5 & \underline{58.8} & 79.9 \\
\tipsvtwo g/14 (ours) & \textbf{75.7} & 95.1 & \textbf{68.9} & \textbf{60.7} & \underline{85.9} & \textbf{72.8} & 80.7 \\
\end{tabular}
}
\vspace{-2mm}
\label{tab:sota_image_text}
\end{table*}

%% file: tables/tab_sota_image_only.tex
\begin{table*}[ht]
\caption{\small\textbf{Image-only evaluations}, where \tipsvtwo achieves the best or second-best performance in $7$ out of $9$ evaluations. We highlight the \textbf{best} and \underline{second-best} number of each column.
}
\vspace{-2mm}
\scriptsize
\addtolength{\tabcolsep}{-0.1em}
\centering
{
\begin{tabular}{l c c c c c c c c c }
\small \multirow{2}{*}{Method} & \multicolumn{2}{c}{Segmentation $\uparrow$} & \multicolumn{2}{c}{Depth $\downarrow$} &
\multicolumn{2}{c}{Normals $\downarrow$} & Fine-grained $\uparrow$ &
\multicolumn{2}{c}{ImageNet Class. $\uparrow$} \\
& PASCAL & ADE20k & NYUv2 & NAVI & NYUv2 & NAVI & retrieval (UnED) & KNN & lin \\
\midrule
CLIP L/14 & 74.5 & 39.0 & 0.553 & 0.073 & 24.3 & 25.5 & 57.4 & 79.8 & 84.3 \\
SigLIP2 SO/14 & 78.1 & 45.4 & 0.466 & 0.064 & 23.0 & 25.0 & - & - & - \\
DINOv2 g/14 & 83.1 & 49.5 & 0.372 & \bf{0.054} & \bf{20.7} & \bf{24.0} & 62.7 & 83.6 & \underline{87.3} \\
PE-core G/14 & - & 41.5 & - & - & - & - & - & \bf{86.8} & \bf{89.5} \\
PE-spatial G/14 & - & 49.3 & - & - & - & - & - & - & - \\
FRANCA G/14 & 81.3 & 42.4 & - & - & - & - & - & 83.0 & 85.9 \\
TIPS g/14 & \underline{83.6} & \underline{49.9} & \underline{0.353} & \underline{0.058} & 21.9 & 24.2 & \bf{68.2} & 83.3 & 86.2 \\
\tipsvtwo g/14 (ours) & \bf{85.1} & \bf{51.6} & \bf{0.334} & 0.059 & \underline{21.7} & \underline{24.1} & \underline{67.0} & \underline{83.7} & 86.8 \\
\end{tabular}
}
\vspace{-2mm}
\label{tab:sota_image_only}
\end{table*}

%% file: tables/tab_dinov3.tex
\begin{table*}[t!]
\caption{\small\textbf{Comparison between DINOv3 and \tipsvtwo}, on largest comparable model size ViT-L. \tipsvtwo achieves superior performance on 4 out of 6 metrics. We highlight the \textbf{best} performing model for each task.}
\vspace{-2mm}
\scriptsize
\addtolength{\tabcolsep}{0.5em}
\centering
{
\begin{tabular}{l c c c c c c}
\small \multirow{2}{*}{Model} & Segmentation $\uparrow$ & Depth $\downarrow$ & 0-Shot $\uparrow$ & I$\rightarrow$T Ret. $\uparrow$ & T$\rightarrow$I Ret. $\uparrow$ & Zero-shot Seg. $\uparrow$ \\
& ADE20k & NYUv2 & ImageNet & COCO & COCO & ADE150 \\
\midrule
DINOv3 L/16 & \textbf{54.9} & 0.352 & \textbf{82.3} & 63.7 & 45.6 & 24.7 \\
\tipsvtwo L/14 (ours) & 51.4 & \textbf{0.339} & 79.7 & \textbf{73.5} & \textbf{57.4} & \textbf{25.1} \\
\end{tabular}
}
\vspace{-2mm}
\label{tab:dinov3}
\end{table*}

%% file: sec/6_conclusion.tex
\newcommand{\showviz}[2]{%
  \resizebox{#1\linewidth}{!}{%
	\setlength{\fboxsep}{0pt}
		\includegraphics[height=#1\linewidth]{images/#2.png}
		\hfill
		\includegraphics[height=#1\linewidth]{images/#2_gt.png}
		\hfill
		\includegraphics[height=#1\linewidth]{images/#2_v2.png}
		\hfill
		\includegraphics[height=#1\linewidth]{images/#2_v1.png}
        \hfill
		\includegraphics[height=#1\linewidth]{images/#2_siglip2.png}
}}

\newcommand{\vizwidth}{1.0}
\begin{figure}[tb]
	\centering
	\showviz{\vizwidth}{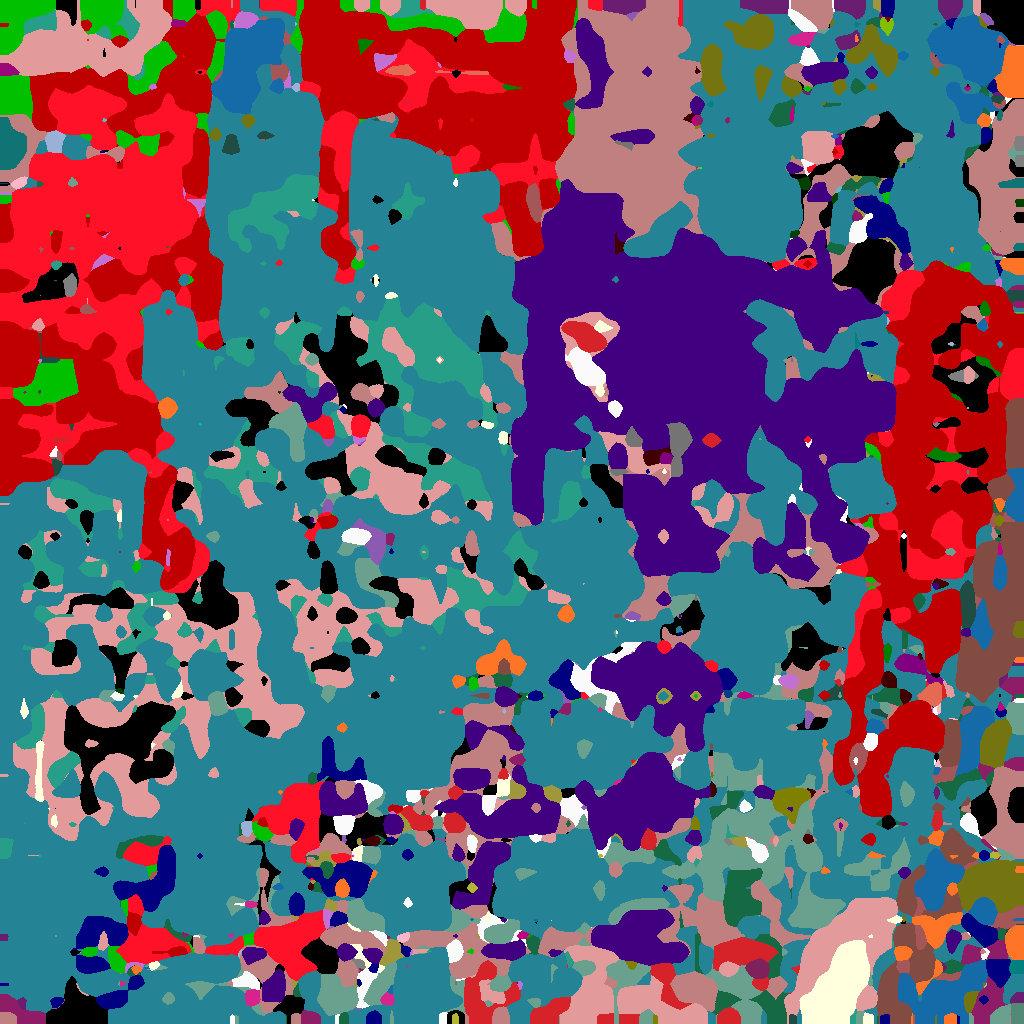} \\[0.2mm]
    \showviz{\vizwidth}{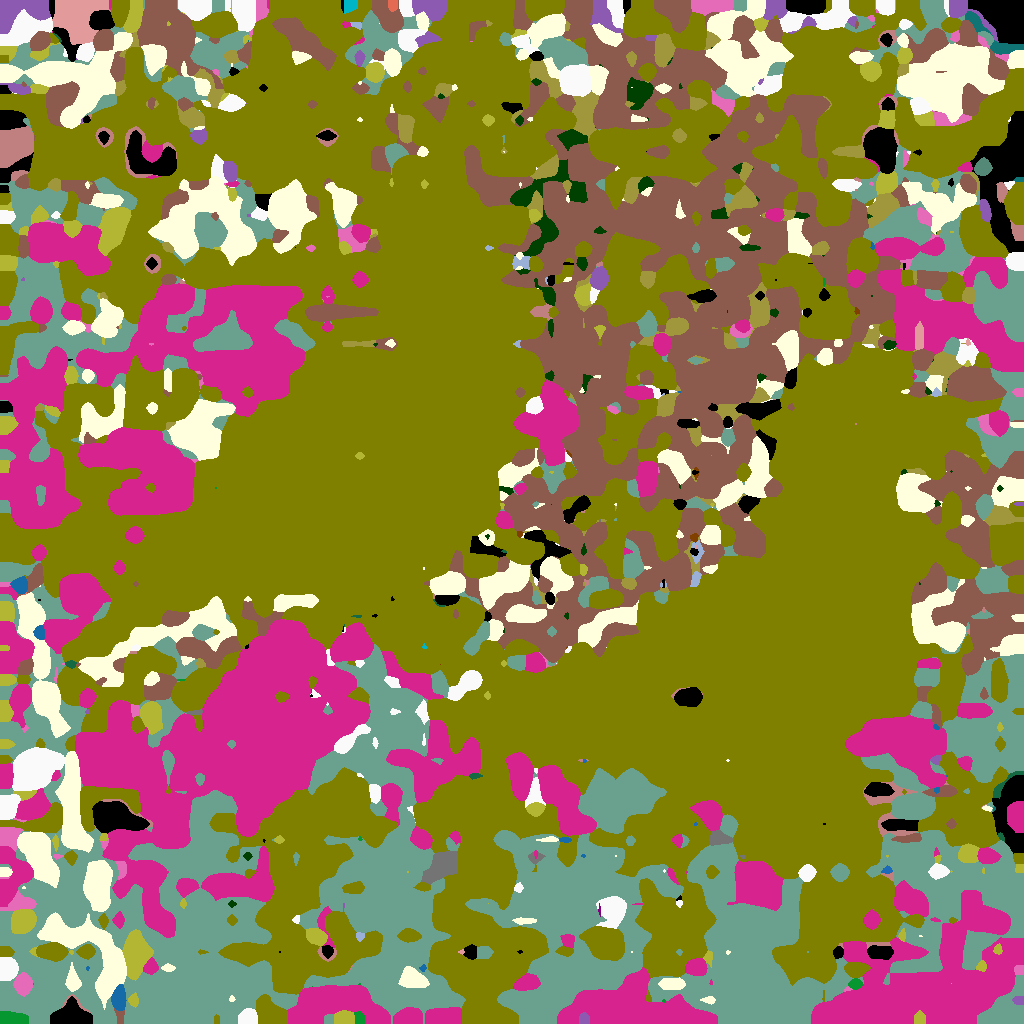} \\[0.2mm]
    \showviz{\vizwidth}{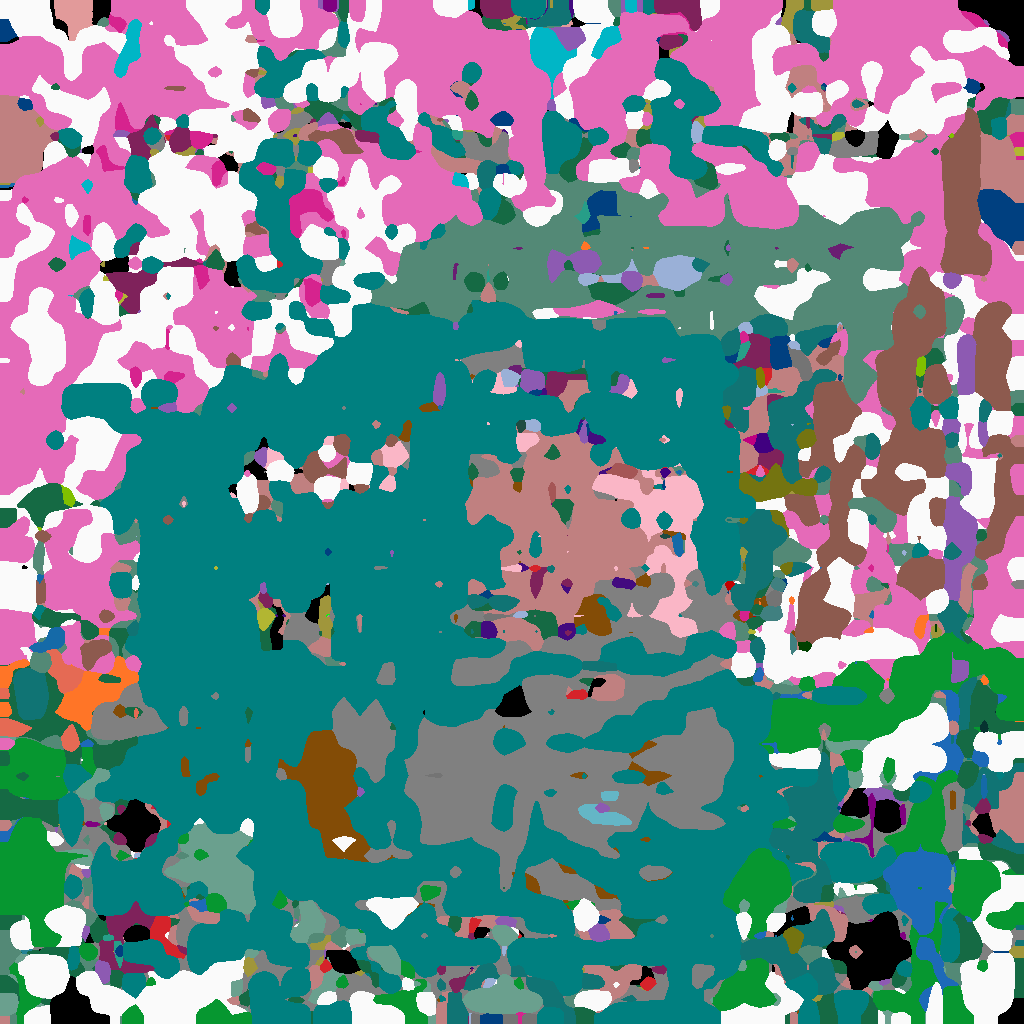} \\[0.2mm]

    \resizebox{\vizwidth\linewidth}{!}{%
    \scriptsize
        \begin{minipage}[t]{0.19\linewidth}
          \centering
          Image
        \end{minipage}%
        \hfill
        \begin{minipage}[t]{0.19\linewidth}
          \centering
          Ground truth
        \end{minipage}%
        \hfill
        \begin{minipage}[t]{0.19\linewidth}
          \centering
          \tipsvtwo (ours)
        \end{minipage}%
        \hfill
        \begin{minipage}[t]{0.19\linewidth}
          \centering
          TIPS
        \end{minipage}%
        \hfill
        \begin{minipage}[t]{0.19\linewidth}
          \centering
          SigLIP2
        \end{minipage}%
    }%
    \vspace{-1mm}
	\caption{\small\textbf{Zero-shot segmentation visualization.} Comparing results for \tipsvtwo, TIPS and SigLIP 2, where classes are predicted directly by finding the closest text embedding to each image patch token, without any post-processing. \tipsvtwo enhances patch-text alignment significantly compared to other models, showcasing strong off-the-shelf capabilities.}
	\label{fig:zss_viz}
	\vspace{-5mm}
\end{figure}

\section{Conclusion}

In this work, we introduce \tipsvtwo, a new vision-language encoder suitable to a variety of multimodal applications.
By investigating the weak dense image-text alignment of previous models, we find that patch-level distillation substantially enhances this capability.
This observation inspires us to introduce iBOT++, an upgrade to the well-known iBOT masked image modeling technique used in pretraining, resulting in significant enhancements to dense image-text alignment.
Additionally, we make vision-language pretraining more efficient with simplifications to the EMA setup, and leverage image captions at different granularities to obtain more robust representations.
Through comprehensive experiments, our final \tipsvtwo model showcases strong results on $9$ tasks spanning $20$ datasets, generally matching or exceeding results from previous vision-language encoders.

\noindent\textbf{Acknowledgments.}~We would like to thank Connor Schenck for thoughtful discussions and suggestions.

%% file: sec/7_supp_mat_for_arxiv.tex
\appendix
\section{Appendix}

\input{tables/tab_clip_full_ema}
\input{tables/tab_clip}
\input{tables/tab_clip_dual}

\subsection{Applying iBOT++ to CLIP}

In the main paper results, we showed the value of our novel iBOT++ loss in the context of the TIPS training recipe.
In this section, we go further to demonstrate the generalizability of iBOT++ to other vision-language architectures, specifically on top of the widely adopted CLIP model.
First, both iBOT and iBOT++ are applied on top of vanilla CLIP, trained with the same dataset used throughout the paper, using a ViT-L backbone.
Our results in \cref{tab:clip_full_ema} demonstrate that integrating iBOT++ yields consistent performance gains across all metrics, beyond that of adding iBOT alone.
A similar conclusion is obtained when using head-only EMA in addition to iBOT++, a setup which reduces memory requirements significantly and achieves generally on par performance with the full EMA case, as reported in \cref{tab:clip}.
Then, as an additional experiment to validate iBOT++, we add it on top of CLIP using two CLS tokens (as in the TIPS setup, but without any of the additional losses used in TIPS), with a ViT-g backbone and the same training dataset (\cref{tab:clip_dual}).
Once again, we see significant and consistent performance gains using iBOT++, compared to the CLIP 2 CLS baseline and its modified version using iBOT.
Notably, in all these cases iBOT++ yields significant improvements in zero-shot segmentation, highlighting its importance for aligning image patches to language.
These results corroborate the usefulness of our proposed iBOT++ recipe for vision-language pretraining.

\subsection{iBOT++ Ablation Study on Masking Ratios}

\input{tables/table_masking_ablation}

As detailed in \cref{bridge}, distillation significantly enhances patch-text alignment, even when the teacher model exhibits poor dense alignment.
In \cref{tab:init}, we identified two key distinctions between the distillation and pretraining stages: A) also applying loss to visible tokens (compare rows (1-2)) and B) lowering masking ratio from  $75\% \rightarrow 0\%$ (compare rows (2-4)) .
In iBOT++, we propose to take the approach in A) from distillation and apply it to pretraining. 
This naturally raises the question: Would applying B) by removing masking in iBOT++ pretraining also further enhance patch-text alignment?
To explore this, we ablated the masking ratio during iBOT++ pretraining. 
Our experiments in \cref{tab:masking} demonstrate that the answer is no: iBOT++ achieves its best performance with a $75\%$ masking ratio.
As a result, we adopt the $75\%$ ratio setting as our final \tipsvtwo training recipe.
Our results confirm that Masked Image Modeling (MIM) remains critical for pretraining, consistently improving performance across image-only tasks and image-text tasks. We conclude that mask removal is only effective during distillation because the teacher model already provides the necessary strong local semantic understanding. This allows the student vision encoder to inherit the alignment without needing to learn it via the MIM objective.

\subsection{Ablations on Multi-Granularity Captions}

In \cref{tab:multi_captions}, we present ablations to justify our strategy to utilize multiple captions, varying 1-2 [CLS] tokens and the assignment to [CLS] tokens between the 3 available captions (web, PaliGemma, Gemini).
These ablations are conducted with the default \tipsvtwo training setup, except that full EMA is used (instead of head-only EMA).
We find the optimal strategy is the recipe in \tipsvtwo: alternating real/synthetic captions with the dual CLS setup.

\input{tables/tab_multicaptions}

\subsection{Qualitative Comparisons to DINOv2 and v3}

In order to further illustrate the comparison with self-supervised pretrained models, we provide additional PCA maps for DINOv2 (with registers)~\cite{vitsneedregisters,oquab2024dinov2} and DINOv3~\citep{siméoni2025dinov3}.
\cref{fig:qualitative_dino_l} presents results on ViT-L models, which is the largest common size between these model families. 
\cref{fig:qualitative_dino_g} presents results on larger models: ViT-g for DINOv2 and \tipsvtwo, but ViT-7B for DINOv3.
Note that the DINOv3 ViT-7B teacher model is trained with 6$\times$ more parameters and 15$\times$ more images than the \tipsvtwo ViT-g teacher.

Comparing DINOv3 and \tipsvtwo, we can see that PCA maps for DINOv3 are smoother, whereas \tipsvtwo maps are more granular.
We note that DINOv3 enhances dense feature maps (on top of DINOv2) by means of a Gram anchoring loss that acts on patch correlations, introducing another teacher that is frozen from the past, along with a higher-resolution inference and down-sampling procedure.
The dense feature maps of \tipsvtwo also demonstrate comparable improvements in spatial coherence (on top of TIPS), but require much simpler changes to the training setup.
On some images, this seems to demonstrate a more semantic focus for \tipsvtwo, compared to a more spatial focus in DINOv3.

For example, in the third row of \cref{fig:qualitative_dino_l} and \cref{fig:qualitative_dino_g}, the backpacks appear semantically similar to the people wearing them in DINOv3 PCA, while they are clearly distinct from the people in \tipsvtwo PCA.
In the second row of \cref{fig:qualitative_dino_g}, the ceiling tends to be underclustered in DINOv3, unable to distinguish lamps and other features; in contrast, \tipsvtwo clearly segments them properly.
In the first row of \cref{fig:qualitative_dino_l}, the windows for DINOv3 PCA are distinctly colored, varying from foreground to background, but are all uniformly colored for \tipsvtwo PCA.

Comparing against DINOv2, \tipsvtwo PCA maps are overall much smoother and more spatially coherent, in contrast to DINOv2's noisy maps.

\newcommand{\showpcas}[2]{%
  \resizebox{#1\linewidth}{!}{%
	\setlength{\fboxsep}{0pt}
		\hfill
		\includegraphics[height=#1\linewidth]{images/supp/#2.jpeg}
		\hfill
		\includegraphics[height=#1\linewidth]{images/supp/dinov2/vit_l/#2.png}
		\hfill
        \includegraphics[height=#1\linewidth]{images/supp/dinov3/vit_l/#2.png}
		\hfill
		\includegraphics[height=#1\linewidth]{images/supp/tipsv2/vit_l/#2.png}
}}

\newcommand{\pcawidthsupp}{1.0}

\begin{figure}[tb]
	\centering
	\showpcas{\pcawidthsupp}{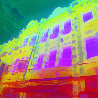} \\[0.2mm]
    \showpcas{\pcawidthsupp}{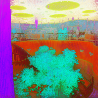} \\[0.2mm]
    \showpcas{\pcawidthsupp}{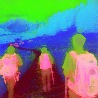} \\[0.2mm]
    \showpcas{\pcawidthsupp}{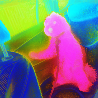} \\[0.1mm]

    \resizebox{\pcawidthsupp\linewidth}{!}{%
    \scriptsize
        \begin{minipage}[t]{0.24\linewidth}
          \centering
          Image
        \end{minipage}%
        \hfill
        \begin{minipage}[t]{0.24\linewidth}
          \centering
          DINOv2-L
        \end{minipage}%
        \hfill
        \begin{minipage}[t]{0.24\linewidth}
          \centering
          DINOv3-L
        \end{minipage}%
        \hfill
        \begin{minipage}[t]{0.24\linewidth}
          \centering
          \tipsvtwo-L (ours)
        \end{minipage}%
    }%
    \vspace{-1mm}
	\caption{\small\textbf{PCA maps at ViT-L size.} Comparing the first $3$ PCA components from the ViT-L models of DINOv2 (with registers), DINOv3, and \tipsvtwo. Images are forwarded at $1372$ resolution for patch size $14$ models (DINOv2 and \tipsvtwo) and at $1568$ resolution for patch size $16$ models (DINOv3). DINOv3 features appear smoother, but \tipsvtwo features show more semantically focused features, e.g., \tipsvtwo maps show all windows clustered together in row 1, and the eyes and leash are distinct on the dog in row 4.}
	\label{fig:qualitative_dino_l}
	\vspace{-3mm}
\end{figure}

\newcommand{\showpcasg}[2]{%
  \resizebox{#1\linewidth}{!}{%
	\setlength{\fboxsep}{0pt}
		\hfill
		\includegraphics[height=#1\linewidth]{images/supp/#2.jpeg}
		\hfill
		\includegraphics[height=#1\linewidth]{images/supp/dinov2/vit_g/#2.png}
		\hfill
        \includegraphics[height=#1\linewidth]{images/supp/dinov3/vit_7b/#2.png}
		\hfill
		\includegraphics[height=#1\linewidth]{images/supp/tipsv2/vit_g/#2.png}
}}

\begin{figure}[tb]
	\centering
	\showpcasg{\pcawidthsupp}{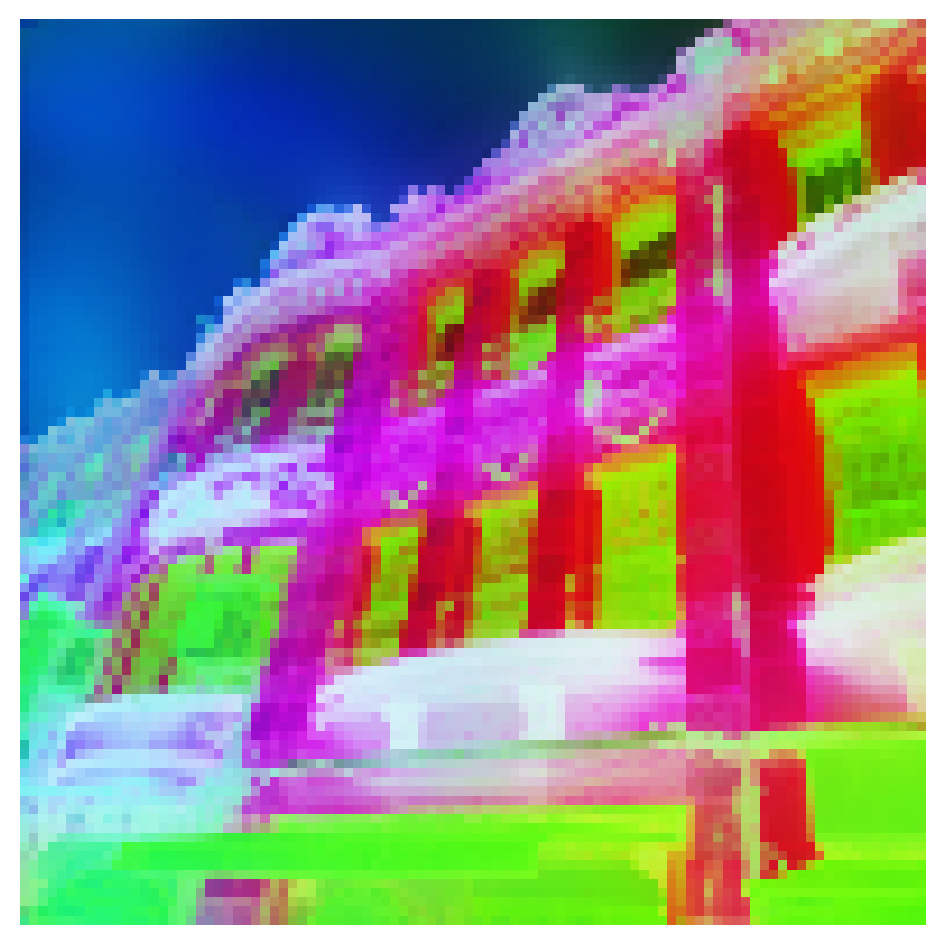} \\[0.2mm]
    \showpcasg{\pcawidthsupp}{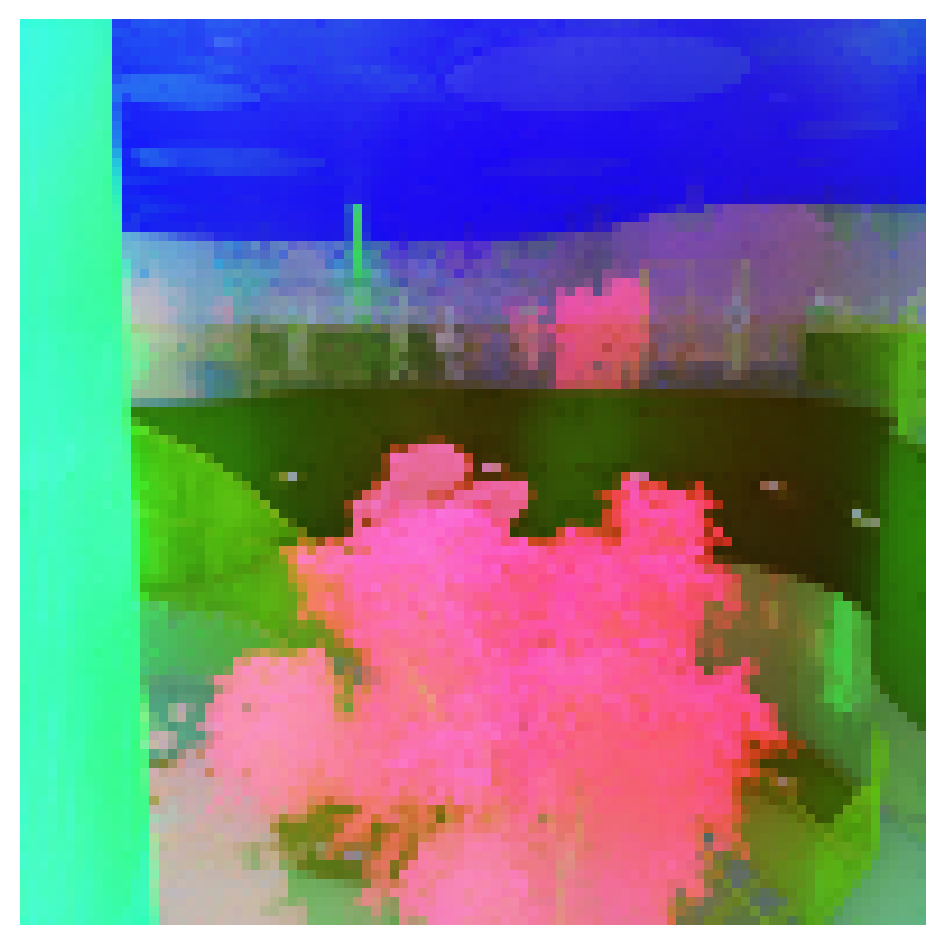} \\[0.2mm]
    \showpcasg{\pcawidthsupp}{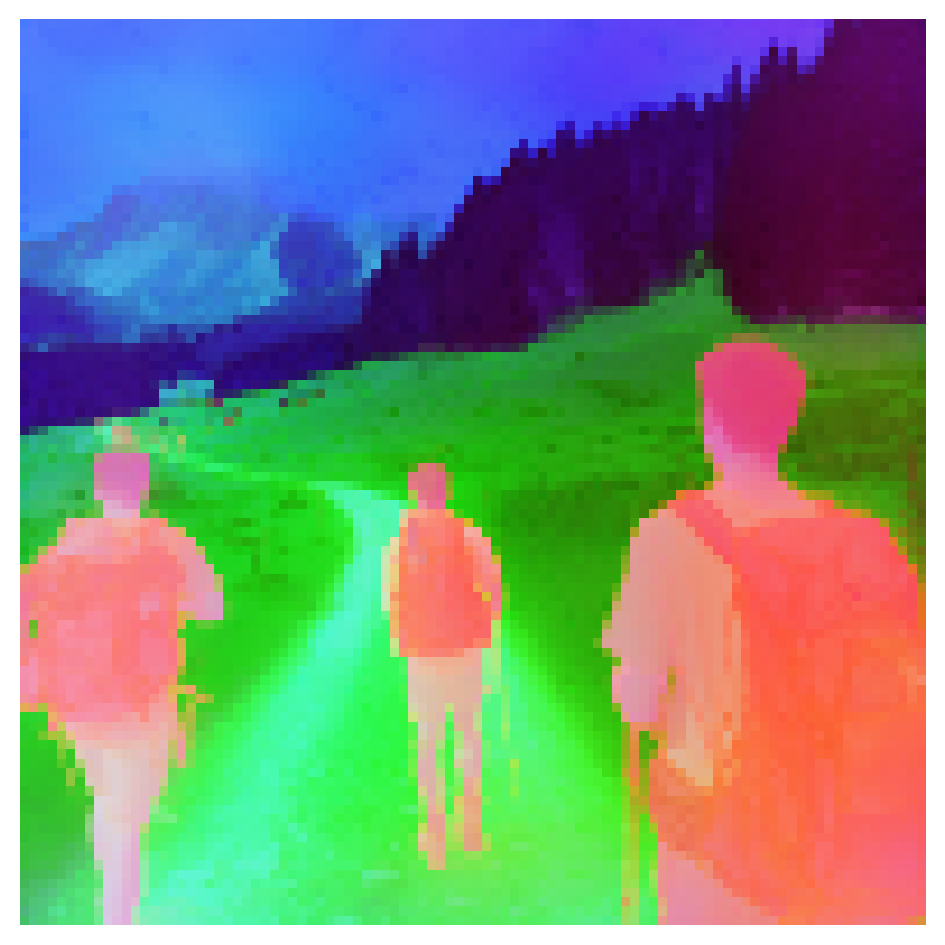} \\[0.2mm]
    \showpcasg{\pcawidthsupp}{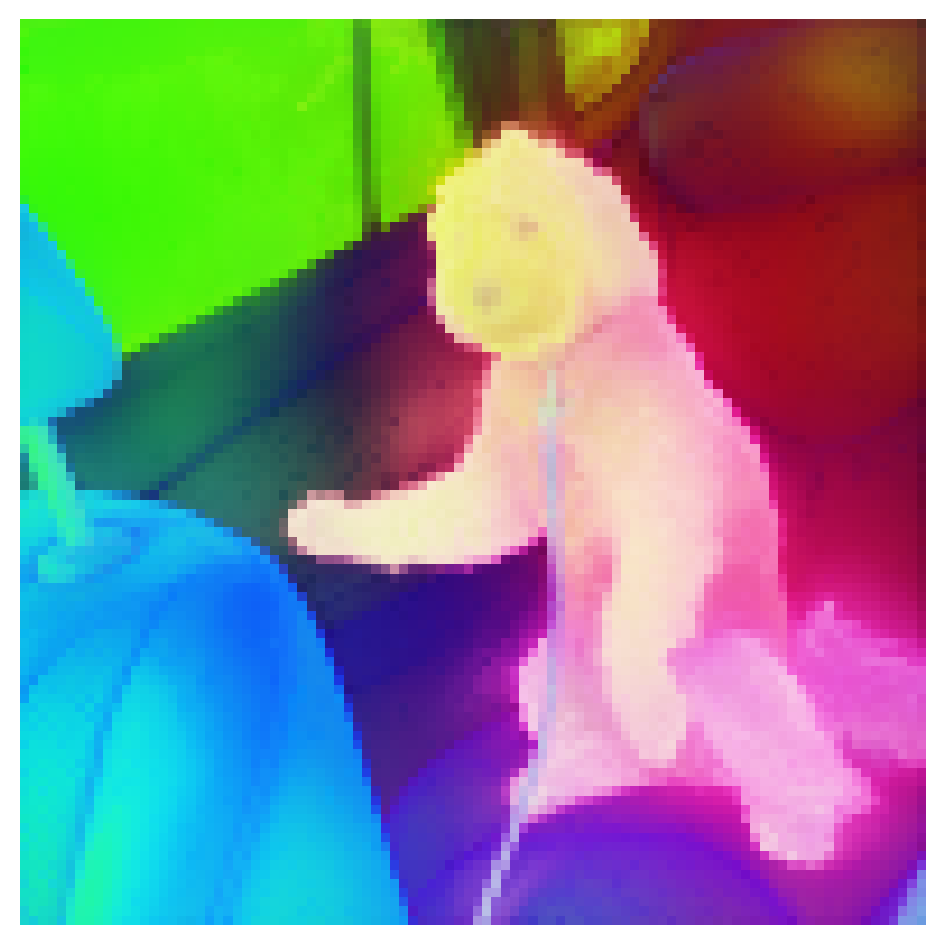} \\[0.1mm]

    \resizebox{\pcawidthsupp\linewidth}{!}{%
    \scriptsize
        \begin{minipage}[t]{0.24\linewidth}
          \centering
          Image
        \end{minipage}%
        \hfill
        \begin{minipage}[t]{0.24\linewidth}
          \centering
          DINOv2-g
        \end{minipage}%
        \hfill
        \begin{minipage}[t]{0.24\linewidth}
          \centering
          DINOv3-7B
        \end{minipage}%
        \hfill
        \begin{minipage}[t]{0.24\linewidth}
          \centering
          \tipsvtwo-g (ours)
        \end{minipage}%
    }%
    \vspace{-1mm}
	\caption{\small\textbf{PCA maps at ViT-g or ViT-7B size.} Comparing the first $3$ PCA components from teacher models of DINOv2  (with registers, ViT-g), DINOv3 (ViT-7B), and \tipsvtwo (ViT-g). Images are forwarded at $1372$ resolution for patch size $14$ models (DINOv2 and \tipsvtwo) and at $1568$ resolution for patch size $16$ models (DINOv3). 
    As for the ViT-L PCA maps, DINOv3 features appear smoother, but \tipsvtwo features capture more semantically focused details, e.g., notice the ceiling details in row 2 or the different colors contrasting people and backpacks in row 3.}
	\label{fig:qualitative_dino_g}
	\vspace{-3mm}
\end{figure}

\subsection{Qualitative Analysis: iBOT++ vs iBOT}

\newcommand{\showvizs}[2]{%
  \resizebox{#1\linewidth}{!}{%
	\setlength{\fboxsep}{0pt}
		\includegraphics[height=#1\linewidth]{images/supp/#2.png}
		\hfill
		\includegraphics[height=#1\linewidth]{images/supp/#2_gt.png}
		\hfill
		\includegraphics[height=#1\linewidth]{images/supp/#2_ibotplus.png}
		\hfill
		\includegraphics[height=#1\linewidth]{images/supp/#2_ibot.png}
}}

\newcommand{\vizwidthsupp}{1.0}
\begin{figure}[tb]
	\centering
	\showvizs{\vizwidthsupp}{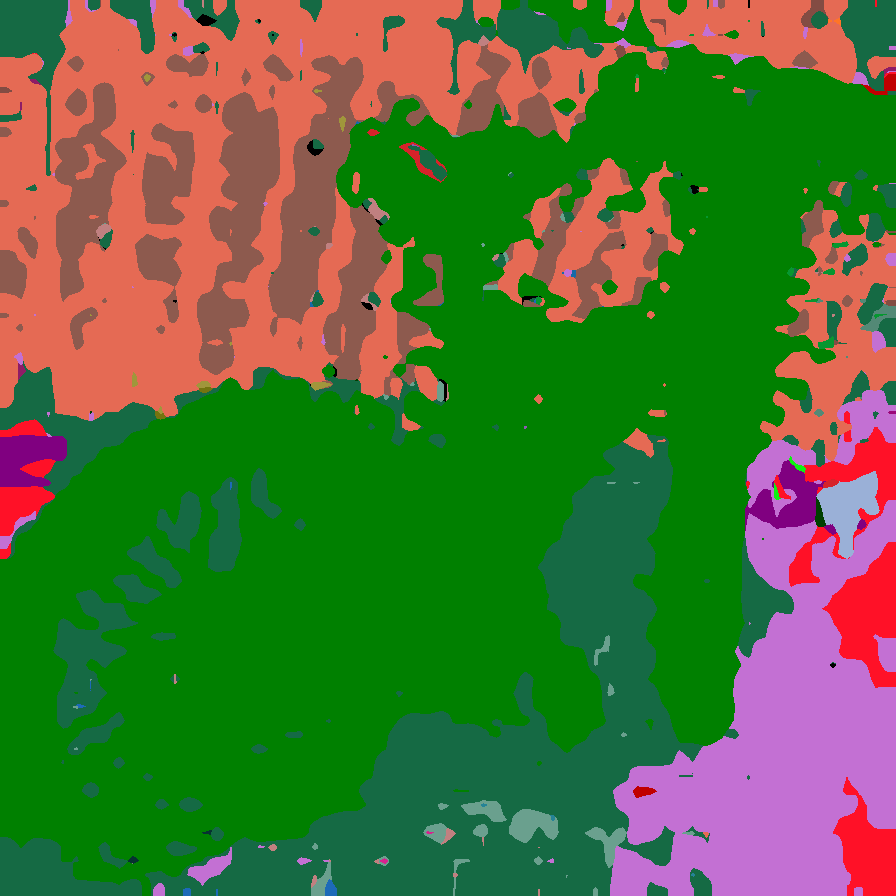} \\[0.24mm]
    \showvizs{\vizwidthsupp}{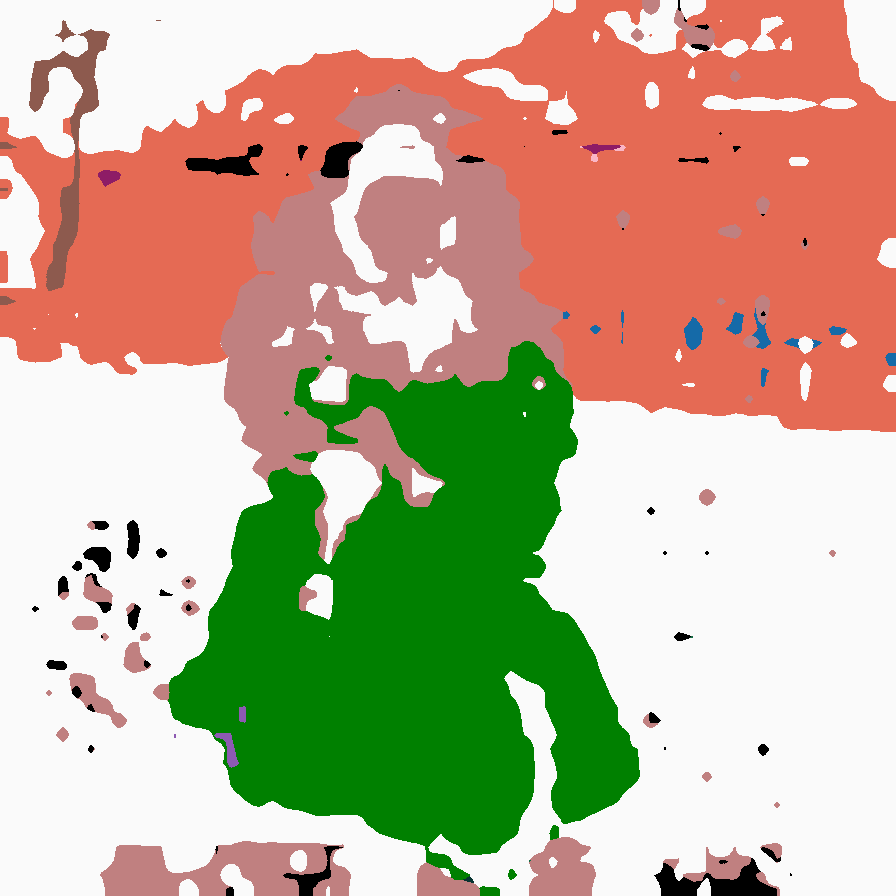} \\[0.24mm]
    \showvizs{\vizwidthsupp}{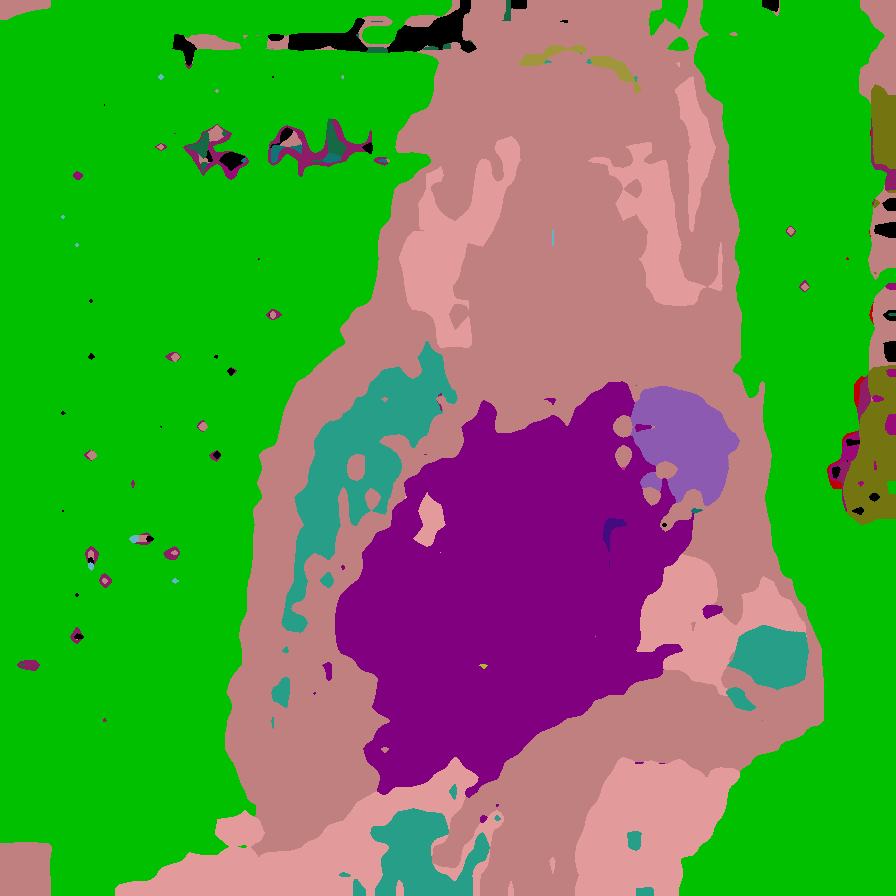} \\[0.24mm]
    \showvizs{\vizwidthsupp}{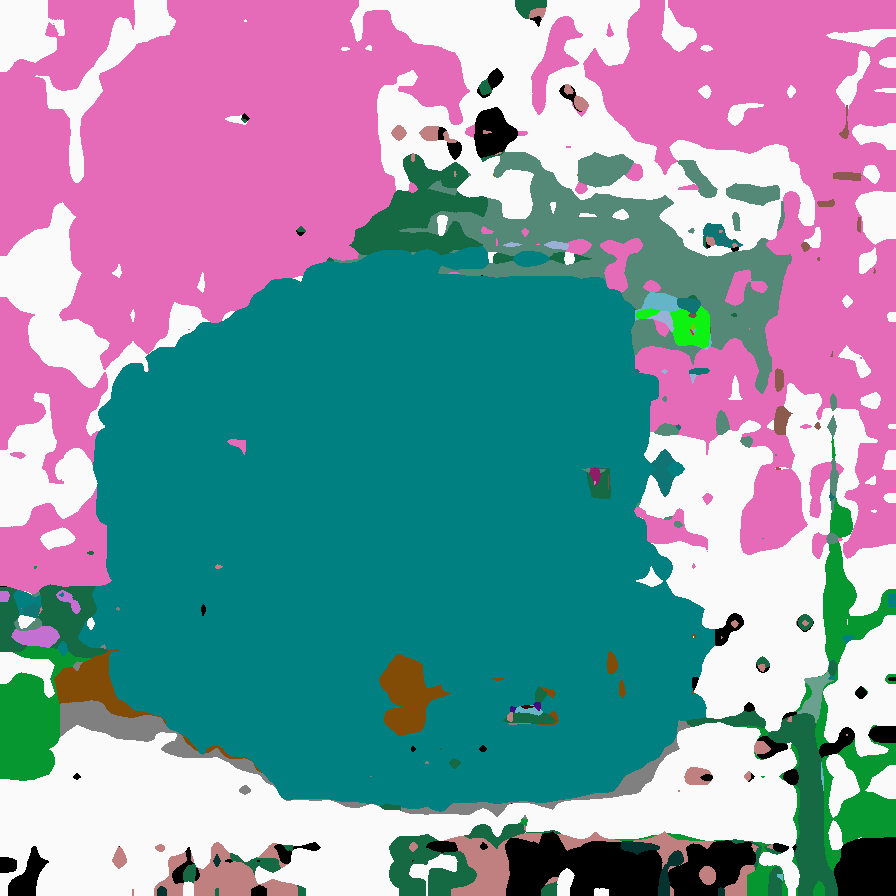} \\[0.24mm]
    \showvizs{\vizwidthsupp}{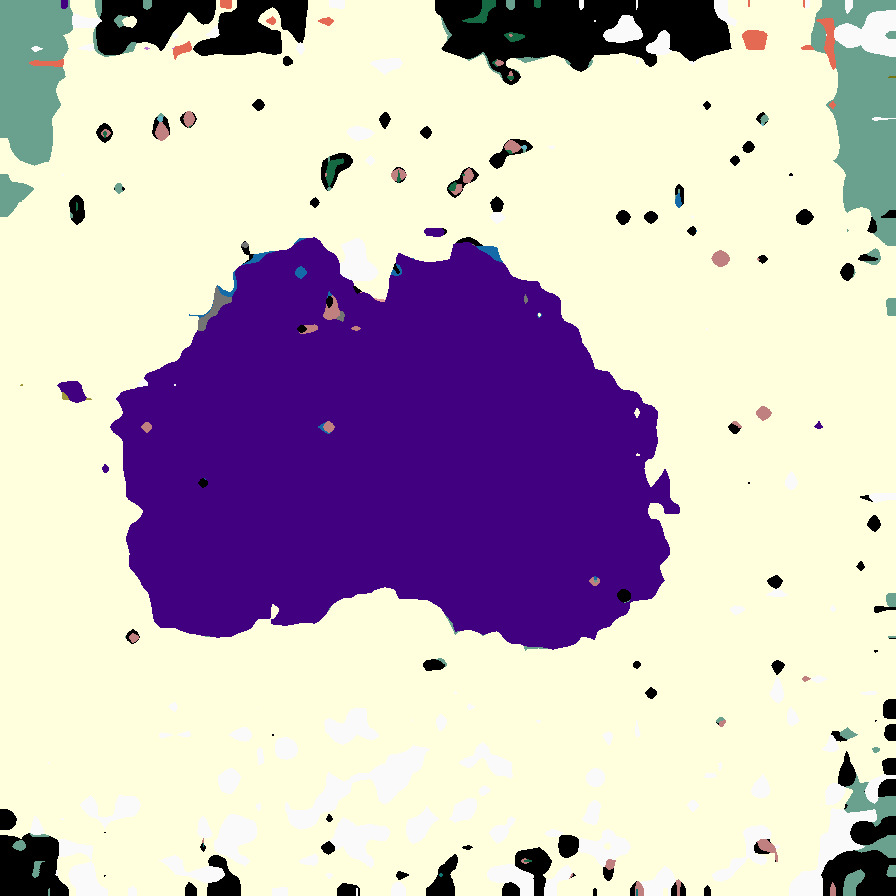} \\[0.24mm]
    \showvizs{\vizwidthsupp}{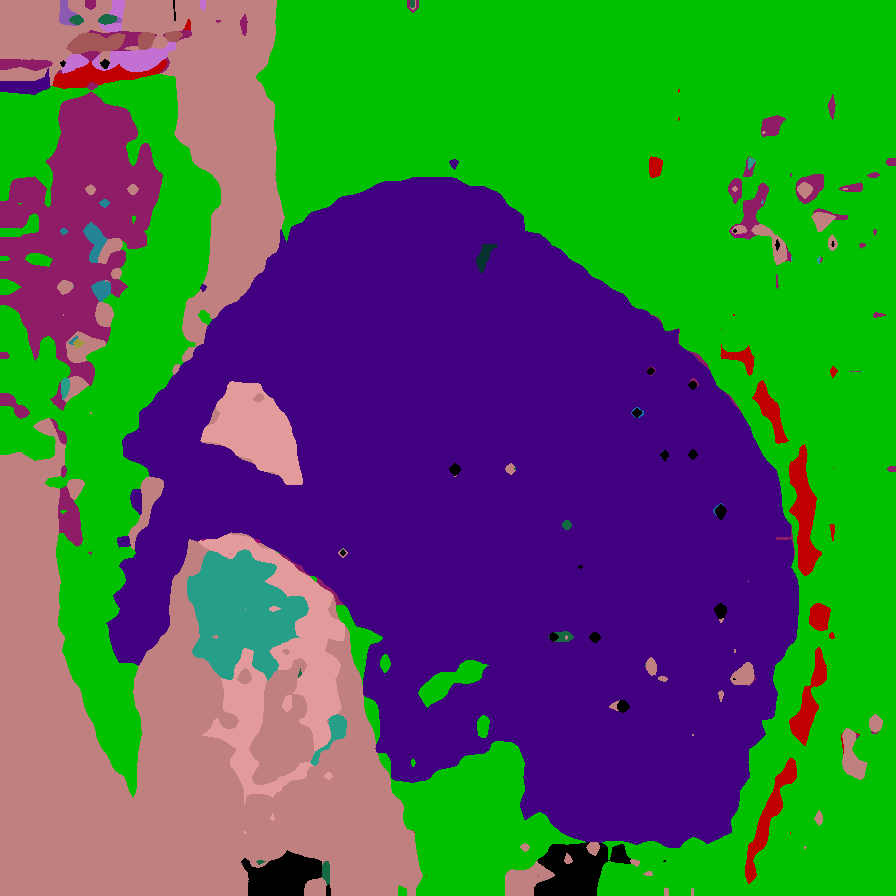} \\[0.24mm]

    \resizebox{\vizwidthsupp\linewidth}{!}{%
    \scriptsize
        \begin{minipage}[t]{0.24\linewidth}
          \centering
          Image
        \end{minipage}%
        \hfill
        \begin{minipage}[t]{0.24\linewidth}
          \centering
          Ground truth
        \end{minipage}%
        \hfill
        \begin{minipage}[t]{0.24\linewidth}
          \centering
          iBOT++
        \end{minipage}%
        \hfill
        \begin{minipage}[t]{0.24\linewidth}
          \centering
          iBOT
        \end{minipage}%
    }%
    \vspace{-1mm}
	\caption{\small\textbf{Zero-shot segmentation visualization.} Comparing results with the iBOT loss (used in TIPS) vs the iBOT++ loss (used in \tipsvtwo), where classes are predicted directly by finding the closest text embedding to each image patch token, without any post-processing. Compared to baseline iBOT, iBOT++ achieves significantly improved patch-text alignment, as part of the TIPSv2 recipe.}
	\label{fig:ibotplus_zss_viz}
	\vspace{-3mm}
\end{figure}

As detailed in \cref{loss_diff}, iBOT++ significantly enhances patch-text alignment during pretraining.
This improvement is quantified in \cref{tab:ablation_studies}, where a substantial performance gain is directly observed in the zero-shot segmentation metrics when comparing iBOT baseline (first row) to iBOT++ (second row). 
To further illustrate this enhancement, \cref{fig:ibotplus_zss_viz} provides qualitative visualizations of the zero-shot segmentation results, which correspond directly to the first and second rows of \cref{tab:ablation_studies}. These visualizations confirm that iBOT++ yields significantly cleaner segmentation maps.

\subsection{Zero-shot Segmentation with SigLIP2}

We showed that smaller models in the TIPS~\cite{tips_paper} family outperform larger ones in the task of zero-shot image segmentation, which motivated our investigations and contributions towards TIPSv2.
In this section, we provide evidence of a similar effect also happening in the SigLIP2 family.
\cref{tab:zeroseg_scale_siglipv2} presents zero-shot segmentation results for three model sizes, where the smallest one outperforms the larger versions in two evaluations, and the SO size model wins in the third; notably, the largest ViT-g model presents the worst performance in all three evaluations.
Note that the smallest ViT-B size model is distilled via active data curation from a larger SO model in the family.
Similarly to the TIPS case, these results indicate that the smaller models surprisingly outperform larger pretrained ones.

\input{tables/tab_zeroseg_scale_siglipv2}
\input{tables/table_family}

\subsection{\textbf{\tipsvtwo} Family} 
\label{sec:panda_family}

The \tipsvtwo model family includes variants with different backbone sizes: ViT-B, ViT-L, ViT-g and SO-400m~\citep{alabdulmohsin2023sovit} (referred to as $\text{SO}$ hereafter). The ViT-g model is directly pretrained, while the smaller ViT-B, ViT-L, and $\text{SO}$ variants are obtained via the patch-level distillation strategy detailed in \cref{sec:preliminaries} and \cref{bridge}, using $\text{ViT}$-g as the teacher model. For text encoder scaling, we follow \citep{tips_paper} and adopt the same transformer parameterization as the image encoder but fix the number of layers at 12 (except for the $\text{SO}$ variant, which maintains its standard layer count). 

Our studies demonstrate that the \tipsvtwo family exhibits strong performance across all image and image-text benchmarks, as summarized in \cref{tab:family}. The parameter count for each model is detailed in \cref{tab:num_params}.

\input{tables/tab_num_params}

\subsection{Performance of \tipsvtwo against Competitors}

\begin{figure}[t]
    \centering
    \includegraphics[width=1.0\linewidth]{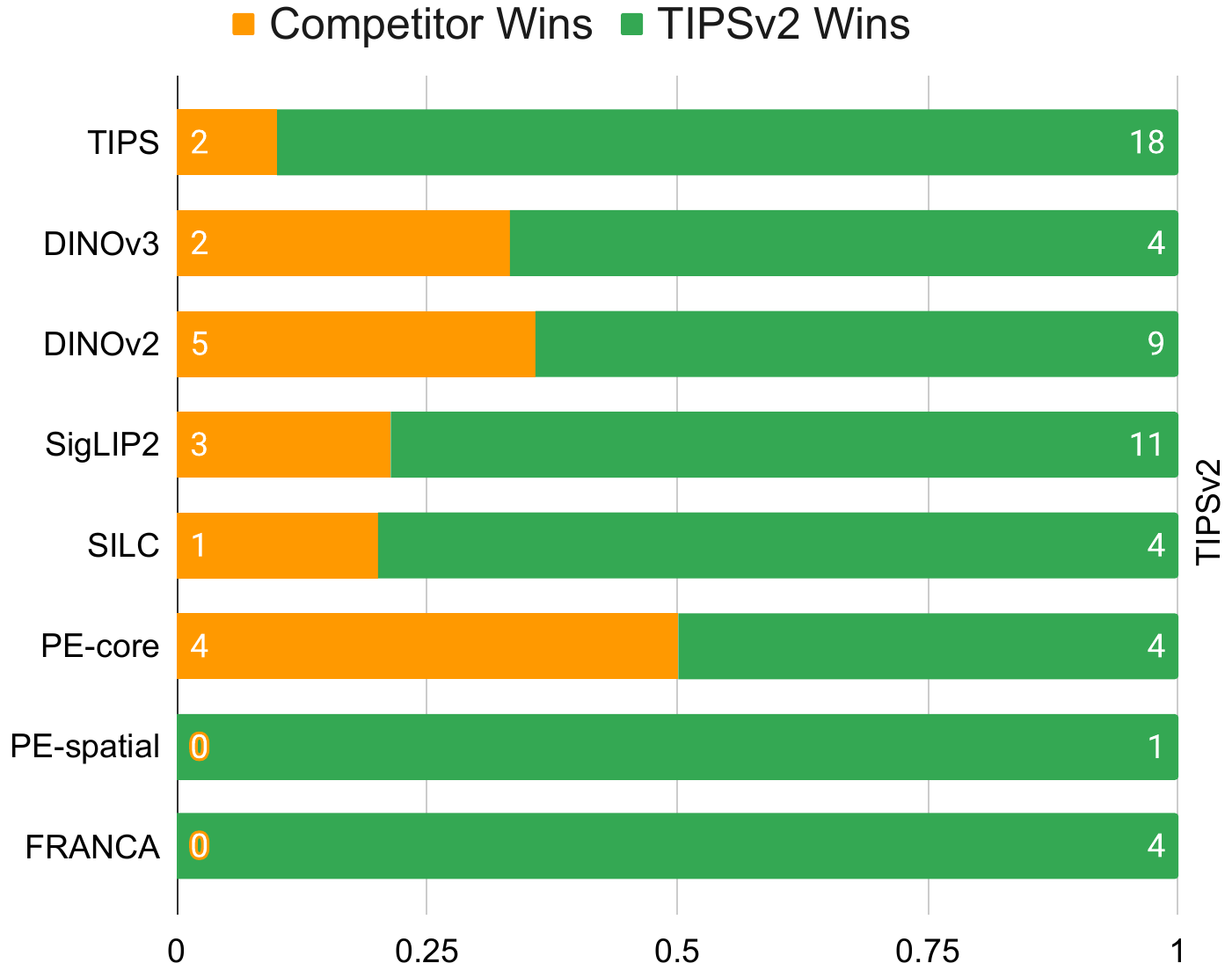}
    \caption{    
    \textbf{Comparison of \tipsvtwo against recent vision encoders.} The chart illustrates the percentage of shared evaluation benchmarks where \tipsvtwo secures the best result (green) compared head-to-head to other individual leading models (orange). 
    \tipsvtwo demonstrates a winning record on the majority of shared tasks. The integer displayed on the chart represents the count of metrics on which each respective model achieves the best result (i.e., the number of wins).
    For each comparison between individual vision encoders, we use the largest comparable model size (ViT-g against ViT-g, or ViT-g against ViT-G, or ViT-L against ViT-L).
    }
    \label{fig:battle}
\end{figure}

To gain a comprehensive understanding of the performance of our new method compared to other recent vision encoders, we quantified the number of evaluations where \tipsvtwo outperforms (or underperforms) each competitor method head-to-head, based on reported scores in the main results tables.
Looking only at the evaluations the models share, \tipsvtwo generally comes out on top in most cases, as shown in \cref{fig:battle}.
This summary confirms the strong results from \tipsvtwo overall.

\subsection{Additional Implementation Details}
 Our implementation closely follows that of TIPS. The loss components are weighted by $\mathcal{L} = \mathcal{L}_\text{CLIP} + \alpha\mathcal{L}_\text{DINO} + \beta\mathcal{L}_\text{iBOT}$, where $\alpha=1.0$, $\beta=2.0$. $\mathcal{L}_\text{CLIP}$ is the averaged result of contrastive losses from the two captions. We use the Adafactor~\citep{shazeer2018adafactor} optimizer, and projection heads are balanced from collapse with EMA centering and sharpening.

%% file: tables/tab_clip_full_ema.tex
\begin{table}[t!]
\caption{\small\textbf{Applying iBOT++ to CLIP}, on a ViT-L backbone. iBOT++ significantly enhances CLIP performance across several tasks, beyond what can be obtained with iBOT.}
\vspace{-2mm}
\scriptsize

\centering

\setlength{\tabcolsep}{3pt} 
\renewcommand{\arraystretch}{1.1}

\resizebox{\columnwidth}{!}{
\begin{tabular}{l c c c c c c}
\small Model & Seg. & Depth $\downarrow$ & ImageNet & IT Ret. & TI Ret. & 0-shot Seg.\\
ViT-L & ADE20k & NYUv2 & KNN & COCO & COCO & PC60 \\
\midrule
CLIP & 35.7 & 0.571 & 73.5 & 52.2 & 33.2 & 4.3 \\
CLIP+iBOT & 41.3 & 0.458 & 77.2 & 55.7 & 37.5 & 8.0 \\
CLIP+iBOT++ & \textbf{42.8} & \textbf{0.434} & \textbf{78.7} & \textbf{58.1} & \textbf{40.6} & \textbf{22.9} \\
\end{tabular}
}
\label{tab:clip_full_ema}
\end{table}

%% file: tables/tab_clip.tex
\begin{table}[t!]
\caption{\small\textbf{Applying iBOT++ to CLIP with head-only EMA}, on a ViT-L backbone. iBOT++ with head-only EMA significantly enhances CLIP performance.}
\vspace{-2mm}
\scriptsize

\centering

\setlength{\tabcolsep}{3pt} 
\renewcommand{\arraystretch}{1.1}

\resizebox{\columnwidth}{!}{
\begin{tabular}{l c c c c c c}
\small Model & Seg. & Depth $\downarrow$ & ImageNet & IT Ret. & TI Ret. & 0-shot Seg.\\
ViT-L & ADE20k & NYUv2 & KNN & COCO & COCO & PC60 \\
\midrule
CLIP & 35.7 & 0.571 & 73.5 & 52.2 & 33.2 & 4.3 \\
CLIP+iBOT & 39.8 & 0.467 & 76.6 & 55.4 & 37.0 & 4.3 \\
CLIP+iBOT++ & \textbf{42.5} & \textbf{0.437} & \textbf{78.2} & \textbf{56.6} & \textbf{39.3} & \textbf{22.1} \\
\end{tabular}
}
\label{tab:clip}
\end{table}

%% file: tables/tab_clip_dual.tex
\begin{table}[t!]
\caption{\small\textbf{Applying iBOT++ to CLIP with a dual CLS setup}, on a ViT-g backbone. iBOT++ significantly enhances performance across several tasks, beyond what can be obtained with iBOT.}
\vspace{-2mm}
\scriptsize

\centering

\setlength{\tabcolsep}{3pt} 
\renewcommand{\arraystretch}{1.1}

\resizebox{\columnwidth}{!}{
\begin{tabular}{l c c c c c c}
\small Model & Seg. & Depth $\downarrow$ & ImageNet & IT Ret. & TI Ret. & 0-shot Seg.\\
ViT-g & ADE20k & NYUv2 & KNN & COCO & COCO & PC60 \\
\midrule
CLIP 2 CLS & 40.1 & 0.475 & 79.1 & 72.9 & 56.1 & 18.2 \\
CLIP 2 CLS + iBOT & 41.1 & 0.442 & 81.6 & 73.2 & 58.3 & 16.4 \\
CLIP 2 CLS + iBOT++ & \textbf{47.2} & \textbf{0.366} & \textbf{82.7} & \textbf{73.6} & \textbf{58.9} & \textbf{28.2} \\
\end{tabular}
}
\label{tab:clip_dual}
\end{table}

%% file: tables/table_masking_ablation.tex
\begin{table}[t!]

\caption{\small \textbf{Masking ablation for iBOT++ pretraining.} We conduct an ablation study on TIPS~\citep{tips_paper} ViT-L models to determine the optimal masking ratio for iBOT++ pretraining. 
The results show that $75\%$ masking ratio is critical for achieving strong performance across all evaluations, particularly enhancing patch-text alignment.
The model achieving the best overall performance is highlighted.}
\vspace{-2mm}
\scriptsize
\setlength{\tabcolsep}{4pt} %
\centering
{
\begin{tabular}{c c c c c c c c}
\small Masking & Segmentation $\uparrow$ & Depth $\downarrow$ & I$\rightarrow$T Ret. $\uparrow$ &  Zero-shot Seg. $\uparrow$ \\
Ratio & PASCAL & NYUv2 & Flickr & ADE150 \\
\midrule
0.0 & 78.8 & 0.513 & 90.0 & 1.0 \\
0.5 & 82.5 & 0.438 & 92.6  & 2.4 \\
0.75 & \textbf{82.6} & \textbf{0.418} & \textbf{93.8} & \textbf{13.6} \\
\end{tabular}
}
\label{tab:masking}
\end{table}

%% file: tables/tab_multicaptions.tex
\begin{table}[t!]
\caption{\small\textbf{Ablations for multi-granularity captions}, on ViT-g. Texts for 2 CLS are separated by \textbf{/} and for each CLS, texts are uniformly sampled from sources. }
\vspace{-3mm}
\centering
\setlength{\tabcolsep}{3pt} 
\renewcommand{\arraystretch}{1.1} 

\resizebox{\columnwidth}{!}{
\begin{tabular}{l c c c c c c}
\multirow{2}{*}{Text Strategy} & Seg. & Depth $\downarrow$ & ImageNet & IT Ret. & 0-shot Seg.  \\
& ADE20k & NYUv2 & KNN & COCO & ADE150 \\
\midrule

1 CLS (web, PaliGemma) & 46.3 & 0.375 & 80.4 & 72.3 & 16.4 \\
1 CLS (web, PaliGemma, Gemini) & 46.7 & 0.366 & 81.2 & 74.4 & 17.7 \\
2 CLS (web / PaliGemma) & 48.3 & 0.370 & \textbf{84.3} & 70.0 & 17.1 \\
2 CLS (web / PaliGemma, Gemini) & \textbf{49.1} & \textbf{0.354} & \textbf{84.3} & 
\textbf{76.2} & \textbf{18.1} \\

\end{tabular}
}
\label{tab:multi_captions}
\end{table}

%% file: tables/tab_zeroseg_scale_siglipv2.tex
\begin{table}[t]
\caption{\small\textbf{Zero-shot segmentation for SigLIP2 models.} Similarly to TIPS, the smaller models tend to outperform larger ones in the same model family.}
\vspace{-2mm}
\scriptsize
\centering
{
\begin{tabular}{l c c c}
\small \multirow{2}{*}{Model} & \multicolumn{3}{c}{Zero-shot Segmentation (mIoU) $\uparrow$} \\
 & PC60 & VOC21 & ADE150 \\
\midrule
SigLIP2 B/16 & \bf{22.6} & 25.8 & \bf{16.4} \\
SigLIP2 SO/14 & 19.6 & \bf{26.8} & 15.6 \\
SigLIP2 g/16 & 17.2 & 25.7 & 13.9 \\
\end{tabular}
}
\vspace{-0mm}
\label{tab:zeroseg_scale_siglipv2}
\end{table}

%% file: tables/table_family.tex
\begin{table*}[t]
\caption{\small \textbf{Evaluations for all \tipsvtwo model variants.} We report four distinct \tipsvtwo model sizes (ViT-B, ViT-L, ViT-g, and SO-400m). Only the ViT-g model is pretrained; the other sizes (ViT-B, ViT-L, and SO-400m) are distilled from the ViT-g teacher. Best performance for each task is highlighted.}
\vspace{-2mm}
\scriptsize
\addtolength{\tabcolsep}{0.5em}
\centering
{
\begin{tabular}{l c c c c c c c}
\small \multirow{2}{*}{Model} & Segmentation $\uparrow$ & Depth $\downarrow$ & Normals $\downarrow$ & KNN $\uparrow$ & I$\rightarrow$T Ret. $\uparrow$ & T$\rightarrow$I Ret. $\uparrow$ & Zero-shot Seg. $\uparrow$ \\
& PASCAL & NYUv2 & NYUv2 & ImageNet & Flickr & Flickr & ADE150 \\
\midrule
\tipsvtwo B/14 & 84.0 & 0.374 & 23.2 & 79.8 & 92.6 & 80.0 & 17.4 \\
\tipsvtwo L/14 & 85.1 & 0.339 & 21.9 & 82.5 & \textbf{95.4} & 83.3 & \textbf{24.7} \\
\tipsvtwo SO/14 & \textbf{85.2} & 0.339 & \textbf{21.7} & 82.8 & 94.8 & 84.0 & 23.3 \\
\tipsvtwo g/14 & 85.1 & \textbf{0.334} & \textbf{21.7} & \textbf{83.7} & 95.1 & \textbf{85.9} & 17.8 \\
\end{tabular}
}
\label{tab:family}
\end{table*}

%% file: tables/tab_num_params.tex
\begin{table}[t]
\scriptsize
\caption{\small\textbf{Number of parameters for all \tipsvtwo model variants.} We release $4$ different model variants. For B, L and g model sizes, we use a fixed number of 12 layers in the text encoder; for the SO size, we use the same number of layers in both the image and text encoders.
}
\addtolength{\tabcolsep}{0.5em}
\centering
{
\begin{tabular}{l c c c}
Model & Image \# Params & Text \# Params &  Total \# Params \\
\midrule
\tipsvtwo-B/14 & 86.3M & 109.6M & 195.9M \\
 \tipsvtwo-L/14 & 304.0M & 183.9M & 487.9M  \\
 \tipsvtwo-SO/14 & 413.3M & 448.3M & 861.7M  \\
 \tipsvtwo-g/14 & 1.1B & 389.1M & 1.5B \\
\end{tabular}
}
\label{tab:num_params}
\end{table}